\documentclass{article}
\usepackage{ijcai21}

\usepackage{times}
\usepackage{soul}
\usepackage{url}
\usepackage[hidelinks]{hyperref}
\usepackage[utf8]{inputenc}
\usepackage[
small
]{caption}
\usepackage{graphicx}
\usepackage{amsmath}
\usepackage{amsthm}
\usepackage{booktabs}
\usepackage{algorithm}
\usepackage{algorithmic}
\usepackage{amsfonts,nicefrac}
\usepackage{multirow,tabularx,diagbox}
\usepackage{microtype}      
\usepackage{dblfloatfix}     
\usepackage{cuted}
\usepackage[switch]{lineno} 
\usepackage{rotating} 
\usepackage{arydshln}
\usepackage{makecell,enumitem}
\usepackage[xcdraw]{xcolor}
\usepackage[font=scriptsize]{subfig}
\usepackage{graphics}

\allowdisplaybreaks

\newtheorem{theorem}{Theorem}
\newtheorem{lemma}{Lemma}

\newcommand{\hide}[1]{}
\newcommand{\JR}[1]{{\color{magenta}JR: #1}}
\newcommand{\reminder}[1]{\textbf{\color{red}[#1]}}  
\newcommand{\ours}{STACK}
\newcommand{\vpara}[1]{\vspace{0.05in}\noindent\textbf{#1 }}

\newcommand{\transpose}[1]{\ensuremath{{#1}^{\scriptscriptstyle T}}}
\renewcommand{\ell}{\mathcal L}
\renewcommand{\theta}{\tau}

\makeatletter
\renewcommand{\maketag@@@}[1]{\hbox{\m@th\normalsize\normalfont#1}}
\makeatother

\title{
Blindfolded Attackers Still Threatening: \\
Strict Black-Box Adversarial Attacks on Graphs
}
\author{
Jiarong Xu$^1$\and
Yizhou Sun$^2$\and
Xin Jiang$^2$\and
Yanhao Wang$^1$\and
Yang Yang$^1$\and
Chunping Wang$^3$\And
Jiangang Lu$^1$
\\
\affiliations
$^1$Zhejiang University\\
$^2$University of California, Los Angeles\\
$^3$FinVolution Group\\
\emails
xujr@zju.edu.cn,
yzsun@cs.ucla.edu,
jiangxjames@ucla.edu,
\{wangyanhao, yangya\}@zju.edu.cn,
wangchunping02@xinye.com,
lujg@zju.edu.cn
}

\begin{document}

\maketitle


\begin{abstract} \label{sec:abstract}

Adversarial attacks on graphs have attracted considerable research interests. Existing works assume the attacker is either (partly) aware of the victim model, or able to send queries to it. These assumptions are, however, unrealistic. To bridge the gap between theoretical graph attacks and real-world scenarios, in this work, we propose a novel and more realistic setting: \textit{strict black-box graph attack}, in which the attacker has no knowledge about the victim model at all and is not allowed to send any queries. To design such an attack strategy, we first propose a \textit{generic graph filter} to unify different families of graph-based models. The strength of attacks can then be quantified by the change in the graph filter before and after attack. By maximizing this change, we are able to find an effective attack strategy, regardless of the underlying model. To solve this optimization problem, we also propose a relaxation technique and approximation theories to reduce the difficulty as well as the computational expense. Experiments demonstrate that, even with no exposure to the model, the Macro-F1 drops 6.4\% in node classification and 29.5\% in graph classification, which is a significant result compared with existent works.
\end{abstract}

\maketitle

\section{Introduction} \label{sec:intro}

Graph-based models, including graph neural networks (GNNs)~\cite{kipf2017semisupervised,velivckovic2017graph} 
and various random walk-based models~\cite{lovasz1993random,perozzi2014deepwalk}, 
have achieved significant success in numerous domains like social network~\cite{pei2015nonnegative}, bioinformatics~\cite{gilmer2017neural}, finance~\cite{paranjape2017motifs}, \emph{etc.}  
However, these approaches have been shown to be vulnerable to adversarial examples~\cite{jiliang2020survey},
namely those that are intentionally crafted to fool the models through slight, even human-imperceptible modifications to the input graph.
Therefore, adversarial attack presents itself as a serious security challenge, 
and is of great importance to identify the weaknesses of graph-based models.

\begin{table}[t]
\centering
\small
\renewcommand\tabcolsep{1pt} 
\begin{tabular}{p{2.3cm}<{\centering}p{1.1cm}<{\centering}p{1.1cm}<{\centering}p{1.3cm}<{\centering}p{1.1cm}<{\centering}p{1.3cm}<{\centering}}
\toprule
Information accessible  &White-box & Gray-box &Restricted black-box & Black-box & \textit{Strict  black-box}  \\\hline
Model parameters & $\surd$ &$\times$ &$\times$ &$\times$ &$\times$\\ 
Labels & $\surd$ & $\surd$ &$\times$ &$\times$ &$\times$ \\ 
Queries & $\surd$ & $\times$ &$\times$ &$\surd$ &$\times$ \\ 
Model structure & $\surd$ & $\bigcirc$  & $\bigcirc$ &$\times$ &$\times$\\ 
\bottomrule
\end{tabular}
\caption{Different graph adversarial attacks, categorized by whether the attacker has full ($\surd$),  limited ($\bigcirc$), or no ($\times$) access to the model.}
\label{tab:attack}
\end{table}

Over recent years, extensive efforts have been devoted to studying adversarial attacks on graphs.  
According to the amount of knowledge accessible to the attacker, 
we summarize five categories of existing works in Table~\ref{tab:attack}. 
As the first attempt, white-box attack assumes the attacker has full access to the victim model~\cite{wu2019adversarial,xu2019topology,chen2018fast,wang2018attack}. 
Then gray-box attack is proposed, in which the attacker trains a surrogate model based on limited knowledge of the victim as well as task-specific labeled data~\cite{zugner2018adversarial,zugner2019adversarial}. 
Whereas, most real-world networks are unlabeled and it is arduously expensive to obtain sufficient labels, which motivates the study of (restricted) black-box attacks. 
With no access to the correct labels, restricted black-box attack shows its success but  still needs limited knowledge of the victim model~\cite{chang2020restricted}.
Comparatively, the black-box model is not aware of the victim model, but has to query some or all examples to gather additional information~\cite{dai2018adversarial,xuan2019unsupervised}. 

Despite these research efforts, there still exists a considerable gap between the existing graph attacks and the real-world scenarios. 
In practice, the attacker cannot fetch even the basic information about the victim model, for example, whether the model is GNN-based or random-walk-based.
Moreover, it is also unrealistic to assume that the attacker can query the victim model in real-world applications.
Such querying attempts will be inevitably noticed and blocked by the defenders.
For example, a credit risk model built upon an online payment network is often hidden behind company API wrappers, and thus totally unavailable to the public.

In view of the above limits, we propose a new attack strategy, \emph{strict black-box graph attack (\ours)}, which has no knowledge of the victim model and no access to queries or labels at all.
The design of such attack strategies is nontrivial due to the following challenges:
\begin{itemize}[leftmargin=*]
\item Most existing graph attack strategies are model-specific, and are not extendable to the strict black-box setting. For example, the attack designed for the low-pass GNN filter~\cite{kipf2017semisupervised,velivckovic2017graph} will inevitable perform badly on a GNN model that covers almost all frequency profiles because the attacker might target at some irrelevant bands of frequency. Thereby, the first challenge is to \textit{identify a common cornerstone for various types of graph-based models.}

    
\item  Existing works always use model predictions or the feedback from surrogate models to quantify the effects of attacks.
However, with no access to the queries and the correct labels, we can neither measure the quality of predictions, nor refer to a surrogate model. So the second challenge is \textit{how to effectively quantify the strength of graph attacks and efficiently compute them.}
\end{itemize}

To handle the above challenges, we first propose a generic graph filter which formulates the key components of various graph models into a compact form.
Then we are able to measure the strength of attacks by the change of the proposed graph filter before and after attack.
An intuitive but provably effective attack strategy is to maximize this change within a fixed amount of perturbation on the original graph.
To solve the optimization problem efficiently, we further relax the objective of this problem to a function of the eigenvalues, and then we show that the relaxed objective can be approximated efficiently via eigensolution approximation theories.
Besides the reduction in computational complexity, this approximation technique also captures inter-dependency between edge flips, which is ignored in previous works~\cite{bojchevski2019adversarial,chang2020restricted}.
In addition, a restart mechanism is applied to prevent accumulation of approximation errors without sacrificing the accuracy.

We summarize our major contributions as follows: 
\begin{itemize}[leftmargin=*]
\item We bring attention to a critical yet overlooked \emph{strict black-box} setting in adversarial attacks on graphs.

\item We propose a generic graph filter applicable to various graph models, and also develop an efficient attack strategy to select adversarial edges.

\item Extensive experiments demonstrate that even when the attacker is unaware of the victim model, the Macro-F1 in node classification drops 6.4\% and that of graph classification drops 29.5\%, both of which are significant results compared with SOTA approaches.
\end{itemize}
The success of our \ours~method breaks the illusion that perfect protection for the victim model could block all kinds of attacks: even when the attacker is totally unaware of the underlying model and the downstream task, it is still able to launch an effective attack.

\section{Strict Black-box Attacks (STACK)} \label{sec:problem} 

Let $G=(V,E)$ be an undirected and unweighted graph with the node set $V = \{v_1, v_2, ..., v_N\}$ and the edge set $E \subseteq  V \times V$.
For simplicity we assume throughout the paper that the graph has no isolated nodes.
The adjacency matrix $A$ of graph $G$ is an $N \times N$ symmetric matrix with elements $A_{ij}=1$ if $\{i,j\} \in E$ or $i=j$, and $A_{ij}=0$ otherwise.
Note that we intentionally set the diagonal elements of $A$ to $1$. 
We also denote by $D$ the diagonal degree matrix with
{\small $D_{ii} = \sum_{i=1}^N A_{ij}$}. 

Suppose we have a graph model designed for certain downstream tasks (\emph{i.e.}, node classification, graph classification, \emph{etc.}) 
The attacker is asked to modify the graph structure and/or node attributes within a fixed budget such that the performance of downstream task degrades as much as possible. 
Here we follow the assumption that the attacker will add or delete a limited number of edges from $G$, resulting in a \textit{perturbed graph} $G^\prime=(V, E^\prime)$.
The setting is conventional in previous works~\cite{chen2018fast,bojchevski2019adversarial,%
chang2020restricted}, and also practical in many scenarios like link spam farms~\cite{gyongyi2005link}, Sybil attacks~\cite{yu2006sybilguard}, \emph{etc}. 

Towards more practical scenarios, we assume the attacker can neither access any knowledge about the victim model nor query any examples. 
We call such attack as the \textit{strict black-box attack (\ours)} on graphs. 
The attacker’s goal is to figure out which set of edges flips can fool various graph-based victim models when they are applied to different downstream tasks.
The generality of the proposed attack model is guaranteed by a generic graph filter, which is introduced in~\S\ref{sec:filer}, and in~\S\ref{sec:optprob} we formulate an optimization problem for construction of such a model.

\subsection{Generic Graph Filter} \label{sec:filer} 
Without access to any information from victim models and queries, we need to take into account the common characteristics of various graph-based victim models when designing the adversarial attack. 
We therefore propose a generic graph filter~$S$:
\begin{equation} \label{eq:s}
    S=D^{-\alpha} A D^{-1+\alpha},  
\end{equation}
where $\alpha \in [0,1]$ is a given parameter to enlarge the filter family. 
Many common choices of graph filters can be considered as special cases of the generic graph filter $S$. 
For instance, when $\alpha=1/2$, the corresponding graph filter $S_\mathrm{sym}=D^{-1/2}AD^{-1/2}$ is symmetric, and used in numerous applications, including spectral graph theory~\cite{chung1997spectral} and the graph convolutional networks (GCNs)~\cite{kipf2017semisupervised,velivckovic2017graph}. 
Another common choice of $\alpha$ is $\alpha=1$, and the resulting graph filter $S_\mathrm{rw}=D^{-1}A$ is widely used in many applications related to random walks on graphs~\cite{chung1997spectral}. 
It is well known that many graph properties (\emph{e.g.}, network connectivity, centrality) can also be expressed in terms of the proposed generic graph filter~\cite{lovasz1993random}. 
Furthermore, it is easy to see the relation between the graph Laplacian ($L=D-A$) and $S$, \emph{i.e.}, $S = I - D^{-\alpha} L D^{-1+\alpha}$, where the change in $S$ and the change in $D^{-\alpha} L D^{-1+\alpha}$ is exactly the same.

The proposed generic graph filters have many interesting properties.
Firstly, its eigenvalues are invariant under isomorphic transformations, \emph{i.e.}, its intrinsic spectral properties (\emph{e.g.}, distances between vertices, graph cuts, the stationary distributions of a Markov chain~\cite{10.1145/3055399.3055463}) are preserved regardless of the choice of $\alpha$.
In addition, the eigenvalues of $S$ are bounded in $[-1, 1]$. 
Hence, the generic graph filter of the original graph $G$ and that of the perturbed graph $G^\prime$ are more comparable from the perspective of graph isomorphism and spectral properties; see~\S\ref{sec:method} for more details.

\subsection{The Optimization Problem} \label{sec:optprob}

With the mere observation of the input graph, our attacker aims to find the set of edge flips that would change the graph filter~$S$ most. 
Considering the dependency structure between the nodes, one adversarial perturbation is easy to propagate to other neighbors via relational information.
Therefore, the generic graph filter $S$ is used instead of the adjacency matrix~$A$, because the impact of one edge flip on $S$ can properly depict such cascading effects on a graph.
Similar ideas are conventional adopted by most graph learning models~\cite{chung1997spectral}. 
To flip the most influential edges, we formulate our attack model as the following optimization problem:
\begin{equation} \label{eq:obj} \small
     \begin{array}{ll}
        \mbox{\normalsize maximize} & \ell_{1}\left(A^{\prime}\right) = \left\|(S^\prime)^k-S^k\right\|_F^2 \\
        \mbox{\normalsize subject to} & A^\prime_{ii} = 1, \quad i=1,\ldots,N \\
        & A^\prime_{ij} \in \{0,1\} \mbox{ for } i \neq j \\
        & \|A^\prime-A\|_0 \leq 2\delta,
    \end{array}
\end{equation}
where the optimization variable is the $N \times N$ symmetric matrix~$A^\prime$. 
The matrix power $S^k$ takes account of all the neighbors \emph{within} $k$ hops (instead of those $k$-hop neighbors) because $A$ is defined to include self-loops.
When targeting on localized kernels (\emph{e.g.}, GCN), we prefer $k=1$ or $2$, while a larger $k$ is recommended to capture higher-order structural patterns.
The last inequality constraint shows the budget $\delta$ with $0 < \delta \ll |E|$, which is consistent with most previous works~\cite{jiliang2020survey,dai2018adversarial}. 
The left part of Figure~\ref{fig:intro} gives an example of the proposed problem. 

\section{Methodology} \label{sec:method} 
\begin{figure}[t]  
	\centering  
	\includegraphics[width=0.48\textwidth]{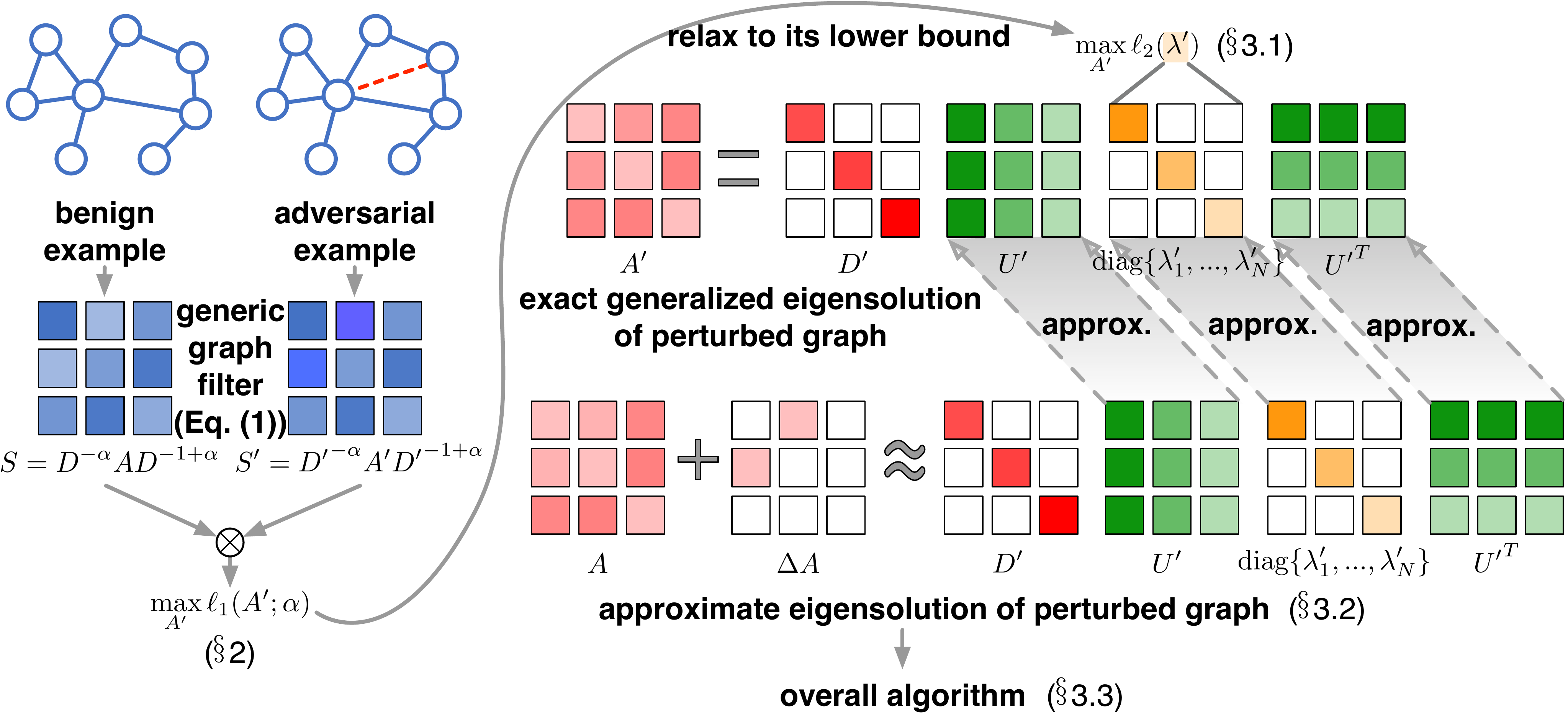}\\  
	\caption{Overview of strict black-box adversarial attacks on graphs.}
    \vspace{-0.4cm}
	\label{fig:intro}     
\end{figure}

Although the objective function $\ell_1(A^\prime)$ described in Problem~\eqref{eq:obj} directly shows the difference between the original graph filter and the perturbed one, the evaluation of the objective function involves a costly eigenvalue decomposition, which in turn makes the whole problem expensive to solve.
Therefore, in this section, we aim to find a sub-optimal solution to Problem~\eqref{eq:obj}, with a reasonable computation expense.
This process is illustrated in Figure~\ref{fig:intro}.
In \S\ref{sec:lbnd} we relax the original objective to a lower bound, which only involves the spectrum of graph filter;
this corresponds to the upper-right part of Figure~\ref{fig:intro}.
Then, we give an approximate solution to compute the perturbed spectrum in~\S\ref{sec:app},
as shown in the lower-right corner of Figure~\ref{fig:intro}.
A detailed description of how to select adversarial edges is provided in~\S\ref{sec:algo} for solving the relaxed problem . 
Finally, in~\S\ref{sec:white-box} we show that our model is flexibly extendable when more information is available.

\subsection{A Viable Lower Bound for $\ell_1$} \label{sec:lbnd}
To find such a relaxation of the objective function $\ell_1(A^\prime)$, we first observe that the eigenvalues of $S$ are independent of $\alpha$, due to the following lemma. 
\begin{lemma} \label{lemma:eigen}
	$\lambda$ is an eigenvalue of {\small$S=D^{-\alpha} AD^{-1+\alpha}$} if and only if {\small$(\lambda, u)$} solves the generalized eigenproblem {\small$Au = \lambda Du$}.
\end{lemma}
With Lemma~\ref{lemma:eigen}, we specify $\alpha=1/2$ and are able to construct a convenient lower bound for $\ell_1$, involving only the eigenvalues of~$S^\prime$.
\begin{theorem} \label{theory:l2}
The function {\small$\ell_1(A^\prime)$} is lower bounded by
\begin{equation} \label{eq:lbnd}\small
    \ell_1(A^\prime) \geq \left(\sqrt{\sum_{i=1}^N (\lambda^\prime_i)^{2k}} - \sqrt{\sum_{i=1}^N  \lambda_i^{2k}}\right)^2 = \ell_2(\lambda^\prime),
\end{equation}
where {\small$\lambda_i$} and {\small$\lambda^\prime_i$} are the {\small$i$}-th
generalized eigenvalue of {\small$A$} and~{\small$A^\prime$}, respectively, i.e., {\small$Au_i=\lambda_iDu_i$} and {\small$A^\prime u^\prime_i=\lambda^\prime_i D^\prime u^\prime_i$}.  We assume that both eigenvalue sequences are numbered in a non-increasing order.
\end{theorem}
Proofs and additional explanation of Lemma~\ref{lemma:eigen} and Theorem~\ref{theory:l2} can be found in~\S\ref{app:proof}.

Now the problem of maximizing $\ell_1 \left(A^{\prime}\right)$ can be properly relaxed to the problem of maximizing its lower bound $\ell_2(\lambda^\prime)$. 
Intuitively, $\ell_2(\lambda^\prime)$ is exactly describing the spectral change, which is a natural measure used in spectral graph theory~\cite{chung1997spectral}.
Additionally, Eq.~\eqref{eq:lbnd} is also valid for any symmetric matrices~$S$ and $S^\prime$; hence, the lower bound $\ell_2(\lambda^\prime)$ holds for different types of networks (\emph{e.g.} weighted or unweighted) and different perturbation scenarios (\emph{e.g.} adjusting edge weights). 

\subsection{Approximating Perturbed Eigensolution} \label{sec:app}

Although $\ell_2(\lambda^\prime)$ is a desirable lower bound for its flexibility, its computation remains a great challenge for two reasons.

First, the computation of $\ell_2(\lambda^\prime)$ relies on the eigenvalue decomposition of~$A^\prime$, which costs $\mathcal O(N^3)$ flops. 
What's worse, this decomposition has to be re-evaluated again and again for every possible set of flipped edges, which turns out to be a computational bottleneck.

Moreover, the inter-dependency between adversarial edges should be taken into consideration. For example, the spectral change caused by flipping two edges $e_1$ and $e_2$ is not identical to the sum of the change caused by flipping~$e_1$ and that caused by~$e_2$.
This difference has been ignored in most studies~\cite{bojchevski2019adversarial,chang2020restricted} but might turn out to be critical in designing adversarial attacks.
In this work, however, we fill this gap by describing such dependency as a combinatorial optimization problem, which is usually difficult to solve.

To resolve the above issues, we can apply the first-order eigenvalue perturbation theory~\cite{stewart1990matrix} to approximate $\ell_2$.
This approximation is accurate enough for our purpose because the perturbation on graphs is assumed to be extremely small, \emph{i.e.}, $\delta \ll |E|$.
We start with the simplest case where only one edge is flipped~\cite{stewart1990matrix}.
\begin{theorem} \label{tm:eigen}
Let {\small$\Delta A = A^{\prime} - A$} and {\small$\Delta D = D^{\prime}-D$} be the perturbations of {\small$A$} and {\small$D$}, respectively. 
Let {\small$(\lambda_k, u_k)$} be the {\small$k$}-th generalized eigen-pair of {\small$A$}, {i.e.}, {\small$Au_k=\lambda_k Du_k$}.
Assume these eigenvectors are properly normalized so that 
{\small$\transpose{u}_j Du_i = 1$} if {\small$i=j$} and {\small$0$} otherwise. 
When only one edge {\small$\{p, q\}$} is flipped, the perturbed {\small$k$}-th generalized eigen-pair can be approximated~by
\begin{linenomath}\small
\begin{align}
	&\lambda^\prime_k \approx \lambda_k + \Delta A_{pq} \big(2u_{kp} \cdot u_{k q}-\lambda_k (u_{kp}^2+u_{kq}^2) \big) \label{eq:eigen-pert-val-2} \\
	& u^{\prime}_k \approx \left( 1- \frac{1}{2}  \Delta A_{p q}  \left(u_{k p}^{2}+u_{k q}^{2}\right) \right) u_k \nonumber \\
	&+\sum_{i \neq k} \frac{\Delta A_{p q} \left( u_{ip}u_{kq}+u_{iq}u_{kp}-\lambda_k \left(u_{ip}u_{kp}+u_{iq}u_{kq}\right) \right) }{\lambda_{ k}-\lambda_{i}} u_{i} \label{eq:eigen-pert-vec-2}
\end{align}
\end{linenomath}
where $u_{kp}$ is the $p$-th entry of the vector $u_k$.
\end{theorem}

For one edge flip, the approximation of the eigenvalue $\lambda_k^\prime$ requires only $\mathcal O(1)$ flops to compute (cf. Eq.~\eqref{eq:eigen-pert-val-2}). 
By ignoring dependency among adversarial edges, some previous works directly adopt Theorem~\ref{tm:eigen} to approximate the spectrum change caused by each edge flip and choose the edges with the highest spectral impact to perturb~\cite{bojchevski2019adversarial,chang2020restricted}. 
To explore and utilize such dependency, we update the eigenvalues and eigenvectors after each edge flip, and then choose the subsequent edge flips based on the updated eigensolutions. 
However, the computational cost associated with re-evaluating the eigenvectors via  Eq.~\eqref{eq:eigen-pert-vec-2}  is still too high ($\mathcal O(N^2)$ flops for each). 
What's worse, Eq.~\eqref{eq:eigen-pert-vec-2} is valid only when all eigenvalues are distinct, which is not guaranteed in practice. 
Therefore, inspired by the power iteration, we propose the following theorem to approximate the perturbed eigenvectors.
\begin{theorem} \label{tm:power}
	Let {\small$\Delta A = A^{\prime} - A$} and {\small$\Delta D = D^\prime - D$} denote the perturbations of  {\small$A$} and {\small$D$}, respectively. Moreover, the number of non-zero entries of  {\small$\Delta A$} and  {\small$\Delta D$}  are assumed to be much smaller than that of {\small$A$} and {\small$D$}. Let {\small$u_k$} be the {\small$k$}-th generalized eigenvector of {\small$A$} with generalized eigenvalue {\small$\lambda_k$}.  We first assume that the eigenvectors are properly normalized, such that {\small$\|u_k\|_2 = 1$}. The $k$-th generalized eigenvector {\small$u^\prime_k$} can then be approximated by
	\begin{equation} \label{eq:power}\small
	    u^\prime_k \approx \begin{cases}
	        \displaystyle  \mathbf{sign} (\lambda_k)  u_k + \frac{\Delta C u_k}{|\lambda_k|}, & \mbox{if } \lambda_k \neq  0 \\
	        \displaystyle \frac{\Delta Cu_k}{\|\Delta Cu_k\|_2}, &\mbox{if } \lambda_k=0.
	    \end{cases}
	\end{equation}
	where {\small$\Delta C=(D+\Delta D)^{-1}(A+\Delta A) - D^{-1}A$}. Specifically, when one edge {\small$\{p,q\}$} is flipped, only the {\small$p$}-th and {\small$q$}-th elements of {\small$u^{\prime}_k $} will be changed because only specific elements  of {\small$\Delta C $} are non-zero, {i.e.},
	{\small
	\[
	\Delta C_{ij} = \begin{cases}
	    A_{pj}^\prime / D_{pp}^\prime - A_{pj} / D_{pp}, \quad &\mbox{if } j \in 	\mathcal N (p) \bigcup \{p,q\} \\
	    A_{qj}^\prime / D_{qq}^\prime - A_{qj} / D_{qq}, &\mbox{if } j \in	\mathcal N (q) \bigcup \{p,q\} \\
	    0, &\mbox{otherwise,}
	\end{cases}
	\]
	where $\mathcal N(p)$ indicates the set of neighbors of node~$p$.
	}
\end{theorem}
When only one edge is flipped, Eq.~\eqref{eq:power} suggests that the approximation of the perturbed eigenvector requires only $\mathcal O(N)$ flops for each, which is far more efficient than $\mathcal O(N^2)$ utilizing Eq.~\eqref{eq:eigen-pert-vec-2}. In addition, the evaluation of $u_k^\prime$ in Eq.~\eqref{eq:power} involves only the $k$-th eigenvalue $\lambda_k$, which removes the strict assumption regarding distinct eigenvalues.

\subsection{Generating Adversarial Edges} \label{sec:algo}

\hide{
\JR{Some previous works directly adopt Theorem~\ref{tm:eigen} to approximate the spectrum change caused by each edge flip and choose the edges with highest spectral impact to perturb~\cite{bojchevski2019adversarial,chang2020restricted}, but they do not consider the dependency among edge flips at all. Considering such dependency when selecting edge flips as a combinatorial optimization problem, we opt to approximately update the eigenvalues and eigenvectors after each edge flip and choose the subsequent edge flips based on the updated eigensolutions.}
}

Up to this point, we have aimed to flip $\delta$ edges so that $\ell_2 \left(\lambda^{\prime}\right)$ is maximized.
Specifically, we first form a candidate set by randomly sampling several edge pairs, as in~\cite{bojchevski2019adversarial}. 
Our attack strategy involves three steps:
(i) for each candidate, compute its impact $\ell_2 \left(\lambda^{\prime}\right)$ on the original graph, such that the eigenvalues can be approximated via Eq.~\eqref{eq:eigen-pert-val-2};
(ii) flip the candidate edge that scores highest on this metric; and (iii) recompute the eigenvalues (Eq.~\eqref{eq:eigen-pert-val-2}) and eigenvectors (Eq.~\eqref{eq:power}) after each time an edge is flipped. 
These steps are repeated until $\delta$ edges have been flipped.

\vpara{Restart mechanism.}
In the above strategy, the approximation of the statistic in step (iii) is efficient. However, it inevitably leads to serious error accumulation as more edges are flipped.
A straightforward solution is to reset the aggregated error (\emph{i.e.}, recompute the exact eigensolutions) over periodic time.
Thus, the question of when to restart should be answered carefully.
Recall that in Theorem~\ref{tm:eigen}, the perturbed eigenvectors must be properly normalized so that $\transpose{u^{\prime}_i} D^{\prime}u^{\prime}_i = 1$.
Thus, we can perform an orthogonality check to verify whether the eigenvectors have been adequately approximated.
Theoretically, if the perturbed eigenvectors are approximately accurate and normalized to $\transpose{u^\prime} D^\prime u^\prime$, then $\transpose{U^{\prime}}D^\prime U^\prime$ will be close to the identity matrix. 
Thus, we propose to use the average of the magnitudes of the non-zero off-diagonal terms of  $\transpose{U^{\prime}}D^\prime U^\prime$ to infer the approximation error of the perturbed eigenvectors:
\begin{equation} \label{eq:error}\small
    \epsilon  = \frac{1}{N(N-1)} \sum_{i=1}^N \sum_{j \neq i} \left|S^\prime_{ij} \right|,
\end{equation}
where $S^\prime = \transpose{U^{\prime}}D^\prime U^\prime$.
The smaller the value of $\epsilon$, the more accurate our approximation.
Thereby, the approximation error~$\epsilon$ is monitored in every iteration, and restart is performed when the error exceeds a threshold $\tau$.
This restart technique ensures approximation accuracy and also reduces time complexity.



\begin{table*}
	\centering	
	\resizebox{2\columnwidth}{!}{
	\renewcommand\tabcolsep{2pt} 
	\begin{tabular}{ccc|cccccccccc|c} 
		\toprule
		&
 &(Unattacked) & \textit{Rand.} & \textit{Deg.} & \textit{Betw.} & \textit{Eigen.} & \textit{DW}  &\textit{GF-Attack} & \textit{GPGD}  &\textit{\ours-r-d} & \textit{\ours-r} & \textit{\ours} &White-box \\ \midrule 
		\multirow{3}{*}{Cora-ML} & {GCN} &0.82$\pm$0.8 & 1.97$\pm$0.8 & 1.12$\pm$0.4 & 1.22$\pm$0.4 & 0.28$\pm$0.3 & 0.85$\pm$0.3 &1.34$\pm$0.5 & 4.22$\pm$0.6 &4.03$\pm$0.6 & 5.02$\pm$0.4 &\textbf{5.27}$\pm$0.3 & 11.36$\pm$0.5 \\
		& {Node2vec} &0.79$\pm$0.8 & 6.37$\pm$1.8 & 5.40$\pm$1.6 & 3.33$\pm$1.0 & 2.84$\pm$1.0 & 3.25$\pm$1.3  &5.76$\pm$1.5 & 5.33$\pm$1.8 &5.82$\pm$1.7 & 6.92$\pm$1.0 & \textbf{8.29}$\pm$1.0 &(1.43$\pm$0.9) \\
		& {Label Prop.}&0.80$\pm$0.7 & 4.10$\pm$1.3 & 2.45$\pm$0.7 & 2.71$\pm$0.8 & 2.07$\pm$0.7 & 1.79$\pm$0.9 &3.18$\pm$0.5 & 4.28$\pm$1.6  &5.01$\pm$0.7 & 6.02$\pm$0.9 & \textbf{7.13}$\pm$0.9 &(1.05$\pm$1.0) \\ \hline
		\multirow{3}{*}{Citeseer} &{GCN} &0.66$\pm$1.4& 2.02$\pm$0.6 & 0.16$\pm$0.4 & 0.70$\pm$0.4 & 0.64$\pm$0.4 & 0.21$\pm$0.4  &1.36$\pm$0.7 & 2.14$\pm$0.9 &2.63$\pm$0.7 & 3.16$\pm$0.6 & \textbf{3.98}$\pm$0.5 &6.42$\pm$0.6\\
		& {Node2vec} &0.60$\pm$1.5& 7.47$\pm$2.3 & 7.47$\pm$1.6 & 3.47$\pm$2.6 & 4.87$\pm$1.5 & 2.54$\pm$2.5 &6.45$\pm$3.5 & 5.26$\pm$1.9  &7.94$\pm$1.6 & 8.32$\pm$2.5 & \textbf{9.32}$\pm$2.6  &(0.12$\pm$1.0)\\
		& {Label Prop.} &0.64$\pm$0.8 &6.70$\pm$2.0  & 3.47$\pm$0.8  & 6.00$\pm$1.7  & 5.36$\pm$0.6  & 3.00$\pm$0.8  &6.99$\pm$1.0 & 5.14$\pm$1.9   &6.66$\pm$1.3 & 7.79$\pm$0.9 & \textbf{8.16}$\pm$0.9  &(2.47$\pm$1.2) \\ \hline
		\multirow{3}{*}{Polblogs} & {GCN} &0.96$\pm$0.7 & 1.91$\pm$1.5 & 0.03$\pm$0.2 & 1.72$\pm$0.6 & 0.67$\pm$0.5 & 0.01$\pm$0.4  &1.15$\pm$0.4 & 2.35$\pm$1.8 &3.06$\pm$1.2 & 4.30$\pm$1.2 & \textbf{5.32}$\pm$1.1 &3.88$\pm$1.1\\
		& {Node2vec} &0.95$\pm$0.3 & 3.01$\pm$0.7  & 0.04$\pm$0.6  & 3.07$\pm$0.6  & 1.84$\pm$0.3  & 0.18$\pm$0.4  &1.00$\pm$0.5  & 2.49$\pm$0.6  &2.57$\pm$0.9 & 2.74$\pm$0.5   & \textbf{3.79}$\pm$0.5  &(2.13$\pm$0.4) \\
		& {Label Prop.} &0.96$\pm$0.5 & 4.99$\pm$0.7 & 0.08$\pm$0.4 & 3.45$\pm$0.7 & 2.15$\pm$0.3 & 0.37$\pm$0.5 &2.18$\pm$0.4 & 4.15$\pm$0.8  &5.17$\pm$0.8 & 5.84$\pm$0.7 &  \textbf{6.14}$\pm$0.7 &(2.28$\pm$0.5)\\ \bottomrule
	\end{tabular}
	}
	\caption{We apply various node-level attacks to different graphs models and different datasets. We report the decrease in Macro-F1 score (in percent) on the test set after the attack is performed; the higher the better. We also report the Macro-F1 on the unattacked graph.}
	\label{tab:node_attack}	\vspace{-1em}
\end{table*}

\subsection{Extension to Different Knowledge Levels} \label{sec:white-box}     

Our proposed adversarial manipulation can also be easily extended to boost performance when additional knowledge (\emph{e.g.}, model structure, learned parameters) is available. One straightforward way of doing this would be adopt $\ell_2 \left(\lambda^{\prime}\right)$ as a regularization term.
Thereby, complementary to other types of adversarial attacks (\emph{e.g.}, white- or gray-box), the proposed attack model 
facilitates rich discrimination on global changes in the spectrum.

As an example of the extension to gray-box attack, assume our goal is to attack a GNN designed for the semi-supervised node classification task.
The attacker is able to alter both the graph structure and the node attributes, and its goal is to misclassify a specific node~$v_i$.
In this case, a surrogate model is built with all non-linear activation function removed~\cite{zugner2018adversarial}; this is denoted as {\small $Z=\mathrm{softmax}(S^kXW)$}, where~$X$ denotes the feature matrix and~$W$ represents the learned parameters.
Therefore, the combined attack model tries to solve the following optimization problem:
\begin{linenomath}\small
\begin{align*}
\mbox{maximize}  \ \max_{c \neq c_0} ( ({S^\prime}^k X^\prime W)_{v_i c} - ({S^\prime}^k XW)_{v_i c_0} )
    + \gamma \ell_2 \left(\lambda^{\prime}\right)
\end{align*}
\end{linenomath}
where the variables are the graph structure $A^\prime$ and the feature matrix $X^\prime$, the scalar $c_0$ denotes the true label or the predicted label for $v_i$ based on the unperturbed graph $G$, while $c$ is the class of $v_i$ to which the surrogate model assigns. The constant $\gamma>0$ is a regularization parameter.
Note that this case can also be considered as an extension to targeted attack. 

\section{Experiments} \label{sec:exp}

In this section, we evaluate the performance of the proposed method on both node classification and graph classification task.
We also extend our model to white-box or targeted attacks, as explained in \S\ref{sec:white-box}. In addition, we study the gap between the initial formulation~\eqref{eq:obj} and the relaxed problem. 

\subsection{Experimental Setup}
\vpara{Datasets.}
For node-level attacks, we adopt three real-world networks for node classification task, Cora-ML, Citeseer, and Polblogs, and we follow the preprocessing in~\cite{dai2018adversarial}.
For graph-level attacks, we use two benchmark protein datasets Enzymes and Proteins for graph classification task. 
See details of each dataset in \S\ref{app:dataset}.

\vpara{Baselines.}
We consider the following baselines.
\begin{itemize}[leftmargin=*]
\item \textit{Rand.}: this method randomly flips edges. 
\item \textit{Deg.}/\textit{Betw.}/\textit{Eigen.}~\cite{bojchevski2019adversarial}: the flipped edges are selected in the decreasing order of the sum of the degrees, betweenness, or eigenvector centrality.
\item \textit{DW}: a black-box attack method designed for DeepWalk~\cite{bojchevski2019adversarial}.
\item \textit{GF-Attack}: targeted attack under the strict black-box setting~\cite{chang2020restricted}. We directly adopt it under our untargeted attack setting by selecting the edge flips on the decreasing order of the loss for SGC/GCN.
\item \textit{GPGD} (Graph Projected Gradient Descent): we apply the GPGD algorithm to solve Problem~\eqref{eq:obj}~\cite{xu2019topology}.
\item \textit{\ours-r-d}: a variant of our method, where both the restart mechanism and the dependency among edge flips are not considered here.
Specifically, the edge flips are selected on the decreasing order according to Eq.~\eqref{eq:eigen-pert-val-2} in Theorem~\ref{tm:eigen}.
\item \textit{\ours-r}: another variant of our method, where the restart mechanism is not considered here.
\end{itemize}

\vpara{Implementation details.}
For the node classification task, we choose GCN~\cite{kipf2017semisupervised}, Node2vec~\cite{grover2016node2vec} and Label Propagation~\cite{zhu2002learning} as the victim models. We set the training/validation/test split ratio as 0.1:0.1:0.8,
and allow the attacker to modify 10\% of the total edges.
Comparatively, GIN~\cite{xu2019how} and Diffpool~\cite{ying2018hierarchical} are used as victim models in the graph classification task due to their excellence. 
We follow the default setting in the above models, including the split ratio, 
and allow the attacker to modify 20\% of the total edges. 

Throughout the experiment, we follow the strict black-box setting.
We set the spatial coefficient $k=1$ and the restart threshold $\tau = 0.03$. 
The candidate set of adversarial edges is randomly sampled in every trial, and its size is set as 20K.
As shown in Table~\ref{tab:node_attack}, this randomness does not hurt the performance overall (Figure~\ref{fig:cand} in \S\ref{app:exp} for more details).
The reported results are all averaged over 10 trials. 
More implementation details can be found in \S\ref{app:imp}. 

\subsection{Experimental Results}

\vpara{Node classification task.}
Table~\ref{tab:node_attack} reports the decrease in Macro-F1 score (in percent) in node classification for all the three datasets and three victim models.
We can see that our method performs the best across all datasets.
First, heuristics (\textit{Rand.}, \textit{Deg.}, \textit{Betw.}, and \textit{Eigen.}) fail to find the most influential edges in strict black-box setting.
Previous works (\textit{DW} and \textit{GF-Attack}) do not perform well either, mainly due to the blindness to additional information (\textit{i.e.}, model type).
The limit of \textit{GPGD} is probably attributed to its relaxation from the binary graph topology to the convex hull.
The ablation study on \textit{Ours-r-d} and \textit{Ours-r} further highlights the significance of the edge dependencies and the restart mechanism.
In addition, the last column in Table~\ref{tab:node_attack} shows the results of white-box attacks which applies \textit{GPGD} for GCN~\cite{xu2019topology}.
The original paper presents great success when using GCN as the victim model.
But our experiments show the failure of its intuitive extension to node-embedding models (\textit{e.g.}, Node2vec) and diffusion models (\textit{e.g.}, Label Propagation).

\begin{table*}[ht!]
\centering
\begin{minipage}{0.7\textwidth}
\resizebox{0.9\textwidth}{!}
{
\renewcommand\tabcolsep{1.2pt} 
\begin{tabular}{cccccccccccc} \toprule
    &  & \textit{Rand.} & \textit{Deg.} & \textit{Betw.} & \textit{Eigen.} & \textit{DW} & \textit{GF-Attack}  & \textit{GPGD}  & \textit{\ours-r-d} & \textit{\ours-r} & \textit{\ours} \\ \midrule
    \multirow{2}{*}{Proteins} & {GIN} & 9.05 & 9.31 & 12.50 & 8.00 & 13.44 &10.82 & 9.49 &11.48 & 12.81 & \textbf{13.53} \\
    & {Diffpool} & 24.13 & 11.27 & 9.87 & 11.03 & 12.71 &21.99 & 14.53 & 23.49 & 24.87 & \textbf{24.88} \\ \hline
    \multirow{2}{*}{Enzymes} & {GIN} & 32.76 & 34.38 & 34.75 & \textbf{40.25} & 38.76 &35.48 & 35.63 &36.00 & 37.46 & 39.90 \\
    & {Diffpool} & 38.09 & 17.48 & 9.51 & 17.74 & 13.55 &37.19 & 20.32  &38.18 & \textbf{40.18} & 39.62 \\ \bottomrule
\end{tabular}
}
\captionsetup{singlelinecheck=false}
\caption{ Graph-level attacks against GIN and Diffpool. We report the decrease in the \\ Macro-F1 score (in percent) on test set. Higher numbers indicate better performance.}
\label{tab:graph_attack}
\end{minipage}%
\begin{minipage}{0.3\textwidth}
\centering
\resizebox{0.9\textwidth}{!}{
\begin{tabular}{lccc} \toprule
	&  \textit{Nettack} & \textit{\ours-ext} \\ \midrule
     {Cora-ML} & 55.70 (0.53) & \textbf{59.75} (0.57) \\
     {Citeseer} &  63.41 (0.60) & \textbf{65.58} (0.63)  \\
     {Polblogs} & 5.16 (0.23)& \textbf{5.40} (0.23) \\
     \bottomrule
\end{tabular}
}
\caption{ Extension to \textit{Nettack}. We report the average decrease in prediction confidence (in percent) of true labels and the misclassification rate in parentheses.}
\label{tab:extension}
\vspace{-1.5em}
\end{minipage}
\end{table*}

\vpara{Graph classification task.}
Table~\ref{tab:graph_attack} reports the results in graph classification.
Especially, our method is 2.65\% on average better than other methods in terms of Macro-F1.
Interestingly, \textit{\ours-r} (without restart) sometimes performs well, 
indicating that we can apply a version with lower complexity in practice yet not sacrificing much accuracy.


\vpara{Extension when additional knowledge is available.}
Here we explore whether our attack can be successfully extended as the example in \S\ref{sec:white-box}.
Specifically, we focus on the \textit{Nettack} model~\cite{zugner2018adversarial} and follow their targeted attack settings.
We randomly select 30 correctly-classified nodes from the test set and mount targeted attacks on these nodes.
From Table~\ref{tab:extension}, we see that the naive extension of our method can further drop 4.84\% in terms of prediction confidence and increase 4.18\% in terms of misclassification rate on average.

\hide{
\begin{table}[t]
\centering
\renewcommand\tabcolsep{3.0pt} 
\small
\resizebox{0.85\columnwidth}{!}{
\begin{tabular}{p{2cm}<{\centering}p{1.8cm}<{\centering}p{1.8cm}<{\centering}p{1.8cm}<{\centering}} \toprule
	& {Cora-ML} & {Citeseer} & {Polblogs} \\ \midrule
	\textit{Nettack} & 55.70 (0.53) & 63.41 (0.60) & 5.16 (0.23) \\
	\textit{\ours-ext} & \textbf{59.75} (0.57)  & \textbf{65.58} (0.63) & \textbf{5.40} (0.23)\\ \bottomrule
\end{tabular}
}
\caption{Extension to \textit{Nettack}. We report the average decrease in prediction confidence (in percent) of true labels and the misclassification rate in parentheses.}
\label{tab:extension}
\end{table}
}

\vpara{Approximation quality.}
From the original objective function~$\ell_1$ to our final model, we have made one relaxation and one approximation; both steps would accumulate errors in evaluating the original objective. 
One thing worth noting is that the estimation error caused by sampling can be ignored here, as demonstrated in~\cite{bojchevski2019adversarial}. 

\begin{figure}[]
    \centering
	\subfloat[Erd\H{o}s--R\'enyi]{
		\includegraphics[width=1.1in]{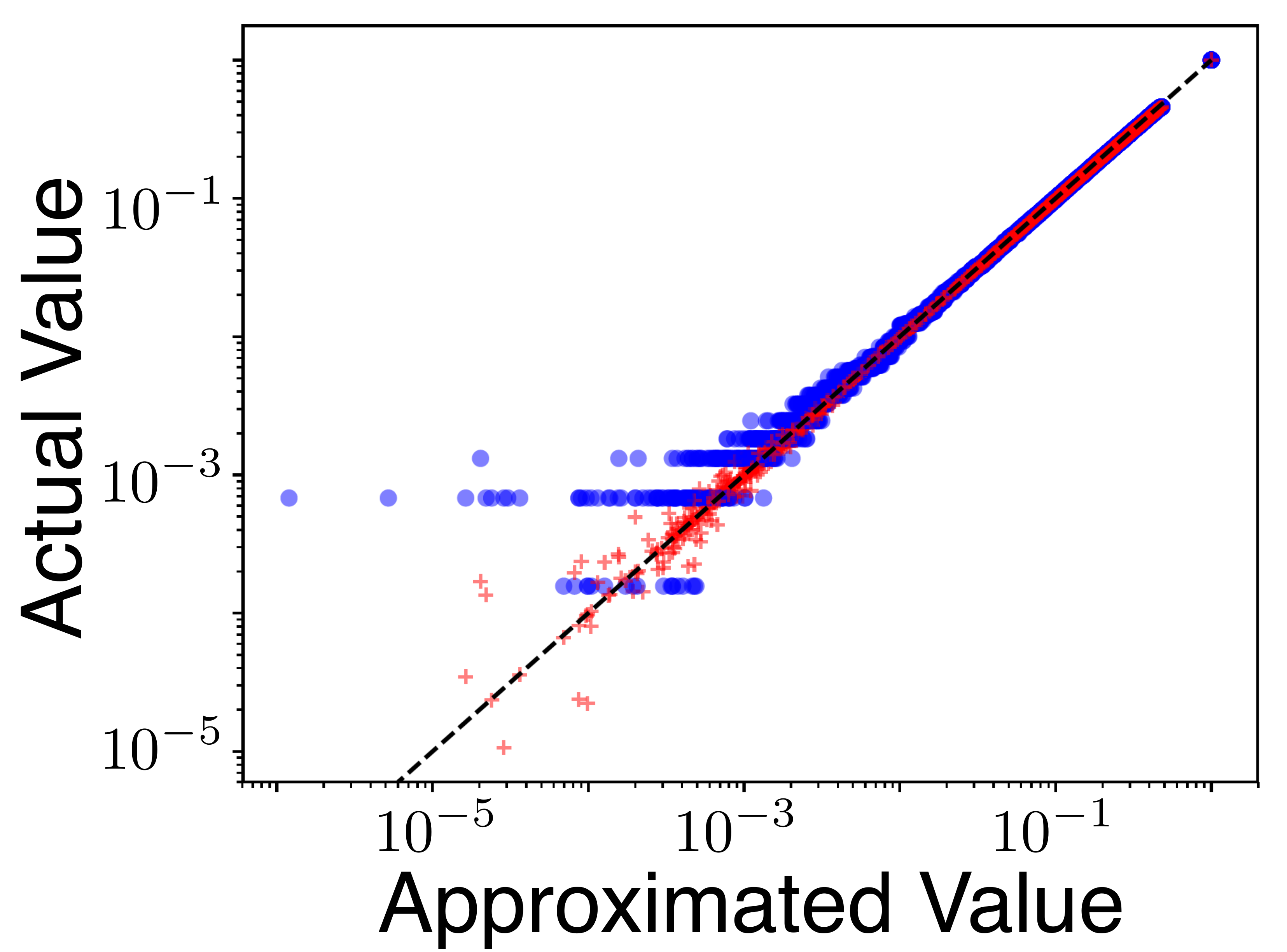}} 
\hspace{0.03\textwidth}
	\subfloat[Barab\'asi--Albert]{
		\includegraphics[width=1.12in]{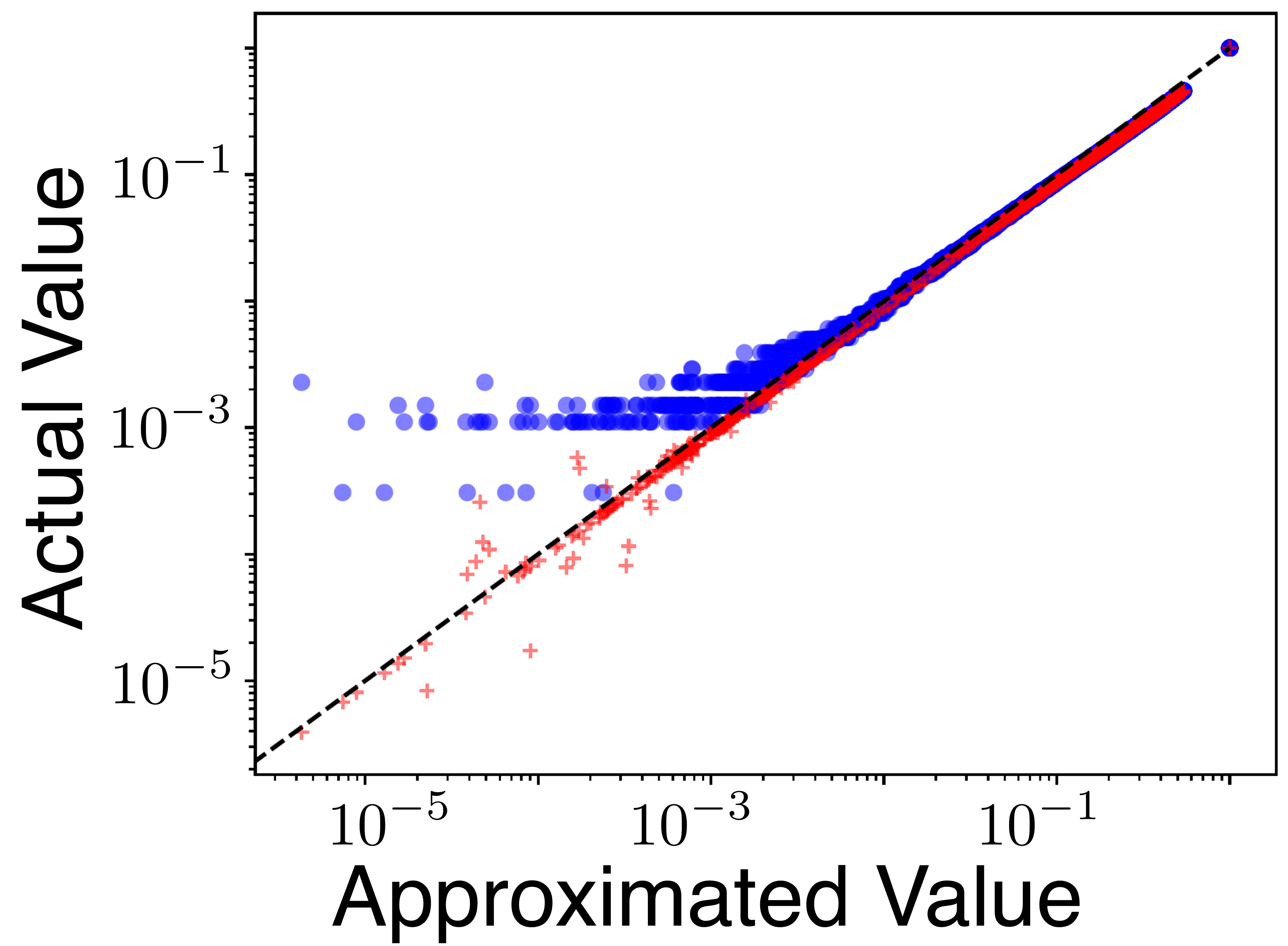}}
	\vspace{-0.01\textwidth}
	\\
	\subfloat[Watts--Strogatz]{
		\includegraphics[width=1.1in]{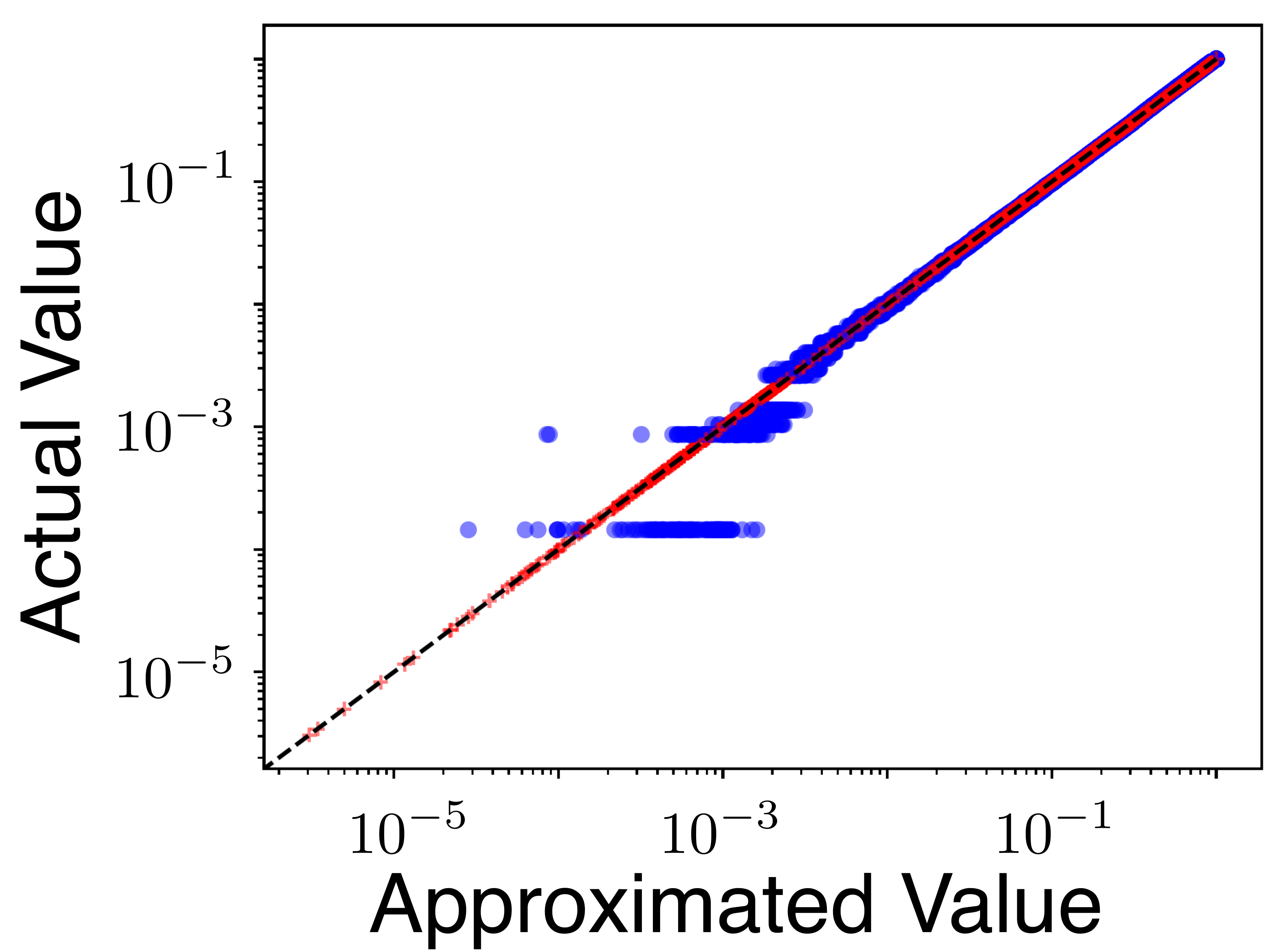}}
\hspace{0.03\textwidth}
	\subfloat[Traid Formation]{
		\includegraphics[width=1.1in]{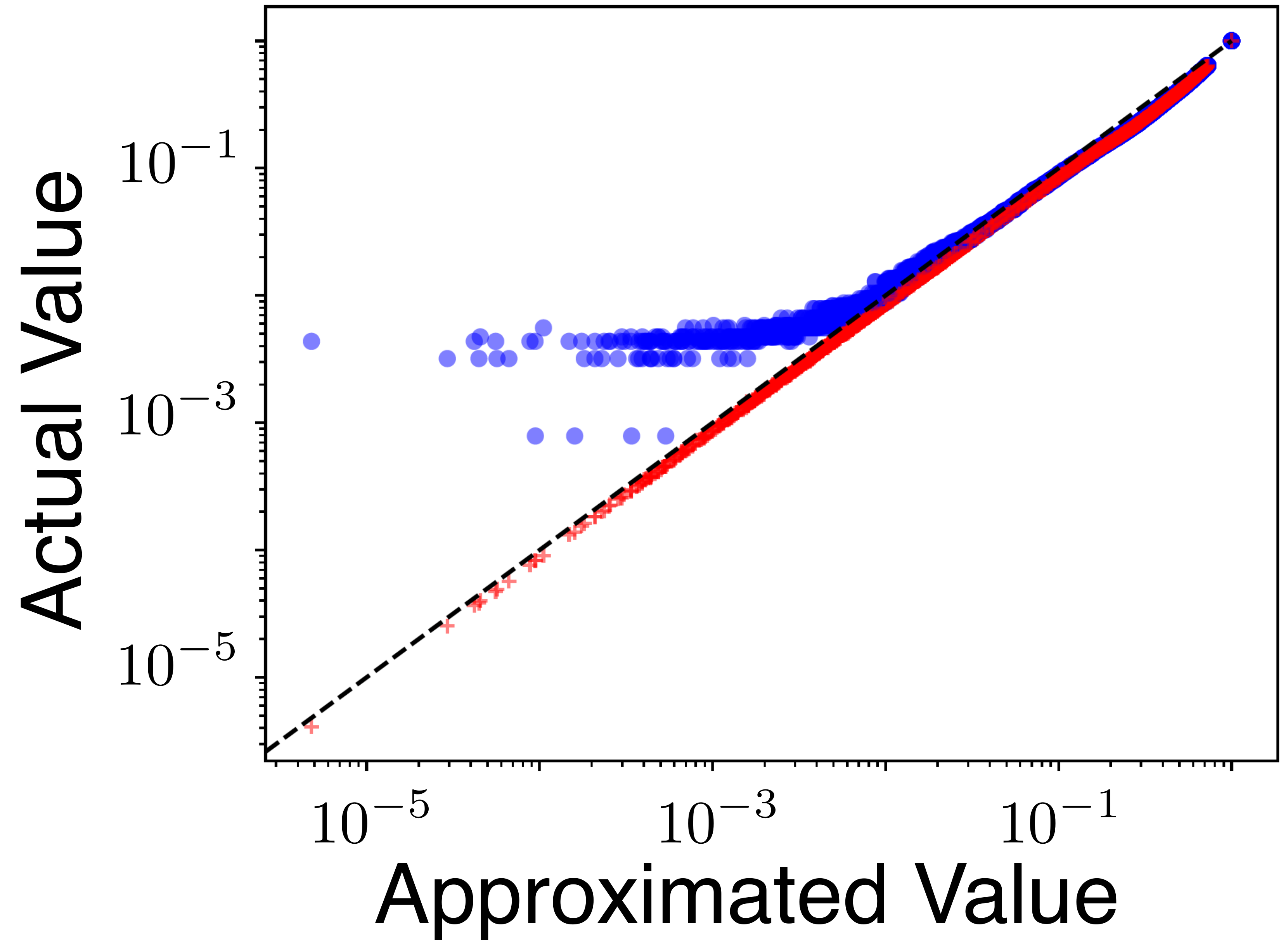}}
	\caption{The true eigenvalues (y-axis) are plotted against the approximations (x-axis) in log scale. We present 
	our approximations with (red plus) and without (blue circle) the restart step. 
	}
	\vspace{-1em}
    \label{fig:approx}
\end{figure}

To 
evaluate the effectiveness of eigen-approximation, 
we first generate the following random graphs with 1K nodes:
Erd\H{o}s--R\'enyi~\cite{bollobas2013modern}, Barab\'asi--Albert~\cite{albert2002statistical}, Watts--Strogatz~\cite{watts1998collective},
and Triad Formation~\cite{holme2002growing}. 
Then we flip 10 edges in every synthetic graph and compare the true eigenvalues with approximated ones. 
Figure~\ref{fig:approx} presents the average results of 100 repeated experiments, where x-axis denotes the approximated value, and y-axis indicates the actual value. 
The approximate linearity of the plotted red points suggests the effectiveness of our eigensolution approximation with restart algorithm (\S\ref{sec:algo}). 
As an ablation study, results generated by our algorithm without the restart step (blue points) shows a tendency of overestimating the change in eigenvalues before and after attacks. 

Furthermore, we study the quality of relaxation and approximation by examining the gap between $\ell_1$ and $\ell_2$.
We compute their Pearson and Spearman correlation coefficients on three real-world datasets 
in Table~\ref{tab:correlation}.
Results are all close to 1, which indicates a linear correlation between the original objective $\ell_1$ and the approximation of $\ell_2$. 

In addition to experiments mentioned in this section, we discuss additional experimental results in \S\ref{app:exp}.

\vspace{-0.3em}
\begin{table}[b]
\centering
\resizebox{0.65\columnwidth}{!}{
\begin{tabular}{cccc} \toprule
    & Cora-ML & Citeseer & Polblogs \\ \midrule
    Pearson & $0.89$ & $0.91$ & $0.85$ \\
    Spearman & $0.91$ & $0.93$ & $0.93$ \\ \bottomrule
\end{tabular}
}
\caption{The Pearson and Spearman correlation coefficients between~$\ell_1$ and approximated $\ell_2$.}
\label{tab:correlation}
\vspace{-1em}
\end{table}


\section{Related Work} \label{sec:related}

We classify the category of graph adversarial attacks into white-box, gray-box, restricted black-box, black-box and strict black-box settings (cf. Table~\ref{tab:attack}).
Specifically, the training of most white-box attacks involve a gradient w.r.t the input of the model~\cite{wu2019adversarial,xu2019topology,chen2018fast,wang2018attack}. 
For gray-box attacks, one common method is to train substitute models as surrogates to estimate the information of the victim models~\cite{zugner2018adversarial,zugner2019adversarial,bojchevski2019adversarial}.
Under restricted black-box attacks, 
\cite{chang2020restricted} assume the family of graph-based models (\emph{i.e.}, GNN-based or sampling-based) is known and design a graph signal processing-based attack method.
For black-box attacks, 
\cite{dai2018adversarial} employ a reinforcement learning-based method and a genetic algorithm to learn from queries of some or all of the examples.
More practically, the strict black-box attacks assume the attacker has totally no knowledge of the victim model and queries.
Besides, another line of work \cite{ma2020towards}, not part of the four categories, assume the victim model is GNN-based but the training input is partly available, which is anther view of practical merit.

\section{Conclusion} \label{sec:conclusion}

In this paper, we describe a strict black-box setting for adversarial attacks on graphs: the attacker not only has zero knowledge about the victim model, but is unable to send any queries as well.
To handle this challenging but more realistic setting, a generic graph filter is proposed to unify different families of graph models;
and its change is used to quantify the strength of attacks.
By maximizing this change, we are always able to find an effective attack strategy.
For efficient solution to the problem, we also propose a relaxation technique and an approximation algorithm. 
Extensive experiments show that the proposed attack strategy substantially outperforms other existing methods.
For future work, we aim to extend STACK to be feasible on node- and attribute-level perturbations.

{\small
\bibliographystyle{named}
\bibliography{reference}

\begin{thebibliography}{}

\bibitem[\protect\citeauthoryear{Albert and
  Barab{\'a}si}{2002}]{albert2002statistical}
R{\'e}ka Albert and Albert-L{\'a}szl{\'o} Barab{\'a}si.
\newblock Statistical mechanics of complex networks.
\newblock {\em Reviews of Modern Physics}, 74:47--97, 2002.

\bibitem[\protect\citeauthoryear{Bojchevski and
  G{\"u}nnemann}{2019}]{bojchevski2019adversarial}
Aleksandar Bojchevski and Stephan G{\"u}nnemann.
\newblock Adversarial {{Attacks}} on {{Node Embeddings}} via {{Graph
  Poisoning}}.
\newblock In {\em ICML}, 2019.

\bibitem[\protect\citeauthoryear{Bollob{\'a}s}{2013}]{bollobas2013modern}
B{\'e}la Bollob{\'a}s.
\newblock {\em Modern graph theory}, volume 184.
\newblock Springer Science and Business Media, 2013.

\bibitem[\protect\citeauthoryear{Chang \bgroup \em et al.\egroup
  }{2020}]{chang2020restricted}
Heng Chang, Yu~Rong, Tingyang Xu, Wenbing Huang, Honglei Zhang, Peng Cui, Wenwu
  Zhu, and Junzhou Huang.
\newblock A restricted black-box adversarial framework towards attacking graph
  embedding models.
\newblock In {\em AAAI}, 2020.

\bibitem[\protect\citeauthoryear{Chen \bgroup \em et al.\egroup
  }{2018}]{chen2018fast}
Jinyin Chen, Yangyang Wu, Xuanheng Xu, Yixian Chen, Haibin Zheng, and Qi~Xuan.
\newblock Fast {{Gradient Attack}} on {{Network Embedding}}.
\newblock {\em ArXiv}, 2018.

\bibitem[\protect\citeauthoryear{Chung}{1997}]{chung1997spectral}
Fan R.~K. Chung.
\newblock {\em Spectral graph theory}.
\newblock American Mathematical Soc., 1997.

\bibitem[\protect\citeauthoryear{Cohen \bgroup \em et al.\egroup
  }{2017}]{10.1145/3055399.3055463}
Michael~B. Cohen, Jonathan Kelner, John Peebles, Richard Peng, Anup~B. Rao,
  Aaron Sidford, and Adrian Vladu.
\newblock Almost-linear-time algorithms for markov chains and new spectral
  primitives for directed graphs.
\newblock In {\em STOC}, 2017.

\bibitem[\protect\citeauthoryear{Dai \bgroup \em et al.\egroup
  }{2018}]{dai2018adversarial}
Hanjun Dai, Hui Li, Tian Tian, Xin Huang, Lin Wang, Jun Zhu, and Le~Song.
\newblock Adversarial {{Attack}} on {{Graph Structured Data}}.
\newblock In {\em ICML}, 2018.

\bibitem[\protect\citeauthoryear{Entezari \bgroup \em et al.\egroup
  }{2020}]{entezari2020all}
Negin Entezari, Saba~A Al-Sayouri, Amirali Darvishzadeh, and Evangelos~E
  Papalexakis.
\newblock All you need is low (rank) defending against adversarial attacks on
  graphs.
\newblock In {\em WSDM}, 2020.

\bibitem[\protect\citeauthoryear{Gilmer \bgroup \em et al.\egroup
  }{2017}]{gilmer2017neural}
Justin Gilmer, Samuel~S. Schoenholz, Patrick~F. Riley, Oriol Vinyals, and
  George~E. Dahl.
\newblock Neural {{Message Passing}} for {{Quantum Chemistry}}.
\newblock In {\em ICML}, 2017.

\bibitem[\protect\citeauthoryear{Grover and
  Leskovec}{2016}]{grover2016node2vec}
Aditya Grover and Jure Leskovec.
\newblock Node2vec: {{Scalable Feature Learning}} for {{Networks}}.
\newblock In {\em SIGKDD}, 2016.

\bibitem[\protect\citeauthoryear{Gyongyi and
  Garcia-Molina}{2005}]{gyongyi2005link}
Zoltan Gyongyi and Hector Garcia-Molina.
\newblock Link spam alliances.
\newblock Technical report, Stanford, 2005.

\bibitem[\protect\citeauthoryear{Holme and Kim}{2002}]{holme2002growing}
Petter Holme and Beom~Jun Kim.
\newblock Growing scale-free networks with tunable clustering.
\newblock {\em Physical Review E}, 65(2):026107, 2002.

\bibitem[\protect\citeauthoryear{Horn and Johnson}{2008}]{horn2008topics}
Roger~A. Horn and Charles~R. Johnson.
\newblock {\em Topics in Matrix Analysis}.
\newblock {Cambridge Univ. Press}, 2008.

\bibitem[\protect\citeauthoryear{Jin \bgroup \em et al.\egroup
  }{2020}]{jiliang2020survey}
Wei Jin, Yaxin Li, Han Xu, Yiqi Wang, and Jiliang Tang.
\newblock Adversarial attacks and defenses on graphs: A review and empirical
  study.
\newblock {\em ArXiv}, 2020.

\bibitem[\protect\citeauthoryear{Kipf and
  Welling}{2017}]{kipf2017semisupervised}
Thomas~N. Kipf and Max Welling.
\newblock Semi-{{Supervised Classification}} with {{Graph Convolutional
  Networks}}.
\newblock In {\em ICLR}, 2017.

\bibitem[\protect\citeauthoryear{Klicpera \bgroup \em et al.\egroup
  }{2019}]{smooth2}
Johannes Klicpera, Aleksandar Bojchevski, and Stephan G{\"u}nnemann.
\newblock Predict then propagate: Graph neural networks meet personalized
  pagerank.
\newblock In {\em ICLR}, 2019.

\bibitem[\protect\citeauthoryear{Li \bgroup \em et al.\egroup }{2018}]{smooth1}
Qimai Li, Zhichao Han, Yuriy Wu, and Xiao-Ming.
\newblock Deeper insights into graph convolutional networks for semi-supervised
  learning,.
\newblock In {\em AAAI}, 2018.

\bibitem[\protect\citeauthoryear{Lov{\'a}sz}{1993}]{lovasz1993random}
L.~Lov{\'a}sz.
\newblock Random walks on graphs: A survey.
\newblock {\em Combinatorics, Paul erdos is eighty}, 2(1):1--46, 1993.

\bibitem[\protect\citeauthoryear{Ma \bgroup \em et al.\egroup
  }{2020}]{ma2020towards}
Jiaqi Ma, Shuangrui Ding, and Qiaozhu Mei.
\newblock Towards more practical adversarial attacks on graph neural networks.
\newblock In {\em NeurIPS}, 2020.

\bibitem[\protect\citeauthoryear{Paranjape \bgroup \em et al.\egroup
  }{2017}]{paranjape2017motifs}
Ashwin Paranjape, Austin~R. Benson, and Jure Leskovec.
\newblock Motifs in {{Temporal Networks}}.
\newblock In {\em WSDM}, 2017.

\bibitem[\protect\citeauthoryear{Pei \bgroup \em et al.\egroup
  }{2015}]{pei2015nonnegative}
Yulong Pei, Nilanjan Chakraborty, and Katia Sycara.
\newblock Nonnegative matrix tri-factorization with graph regularization for
  community detection in social networks.
\newblock In {\em IJCAI}, 2015.

\bibitem[\protect\citeauthoryear{Perozzi \bgroup \em et al.\egroup
  }{2014}]{perozzi2014deepwalk}
Bryan Perozzi, Rami {Al-Rfou}, and Steven Skiena.
\newblock {{DeepWalk}}: Online learning of social representations.
\newblock In {\em SIGKDD}, 2014.

\bibitem[\protect\citeauthoryear{Stewart and Sun}{1990}]{stewart1990matrix}
G.~W. Stewart and Ji-guang Sun.
\newblock {\em Matrix Perturbation Theory}.
\newblock Computer Science and Scientific Computing. {Academic Press},
  {Boston}, 1990.

\bibitem[\protect\citeauthoryear{Veli{\v{c}}kovi{\'c} \bgroup \em et al.\egroup
  }{2018}]{velivckovic2017graph}
Petar Veli{\v{c}}kovi{\'c}, Guillem Cucurull, Arantxa Casanova, Adriana Romero,
  Pietro Lio, and Yoshua Bengio.
\newblock Graph attention networks.
\newblock In {\em ICLR}, 2018.

\bibitem[\protect\citeauthoryear{Wang \bgroup \em et al.\egroup
  }{2018}]{wang2018attack}
Xiaoyun Wang, Joe Eaton, Cho-Jui Hsieh, and Shyhtsun~Felix Wu.
\newblock Attack {{Graph Convolutional Networks}} by {{Adding Fake Nodes}}.
\newblock {\em ArXiv}, 2018.

\bibitem[\protect\citeauthoryear{Watts and
  Strogatz}{1998}]{watts1998collective}
Duncan~J. Watts and Steven~H. Strogatz.
\newblock Collective dynamics of `small-world' networks.
\newblock {\em Nature}, 393(6684):440--442, 1998.

\bibitem[\protect\citeauthoryear{Wu \bgroup \em et al.\egroup
  }{2019}]{wu2019adversarial}
Huijun Wu, Chen Wang, Yuriy Tyshetskiy, Andrew Docherty, Kai Lu, and Liming
  Zhu.
\newblock Adversarial {{Examples}} for {{Graph Data}}: {{Deep Insights}} into
  {{Attack}} and {{Defense}}.
\newblock In {\em IJCAI}, 2019.

\bibitem[\protect\citeauthoryear{Xu \bgroup \em et al.\egroup
  }{2019a}]{xu2019topology}
Kaidi Xu, Hongge Chen, Sijia Liu, Pin-Yu Chen, Tsui-Wei Weng, Mingyi Hong, and
  Xue Lin.
\newblock Topology {{Attack}} and {{Defense}} for {{Graph Neural Networks}}:
  {{An Optimization Perspective}}.
\newblock In {\em IJCAI}, 2019.

\bibitem[\protect\citeauthoryear{Xu \bgroup \em et al.\egroup
  }{2019b}]{xu2019how}
Keyulu Xu, Weihua Hu, Jure Leskovec, and Stefanie Jegelka.
\newblock How {{Powerful}} are {{Graph Neural Networks}}?
\newblock In {\em ICLR}, 2019.

\bibitem[\protect\citeauthoryear{Ying \bgroup \em et al.\egroup
  }{2018}]{ying2018hierarchical}
Zhitao Ying, Jiaxuan You, Christopher Morris, Xiang Ren, William~L. Hamilton,
  and Jure Leskovec.
\newblock Hierarchical {{Graph Representation Learning}} with {{Differentiable
  Pooling}}.
\newblock In {\em NeurIPS}, 2018.

\bibitem[\protect\citeauthoryear{Yu \bgroup \em et al.\egroup
  }{2006}]{yu2006sybilguard}
Haifeng Yu, Michael Kaminsky, Phillip~B Gibbons, and Abraham Flaxman.
\newblock Sybilguard: defending against sybil attacks via social networks.
\newblock In {\em SIGCOMM}, 2006.

\bibitem[\protect\citeauthoryear{Yu \bgroup \em et al.\egroup
  }{2020}]{xuan2019unsupervised}
Shanqing Yu, Jun Zheng, Lihong Chen, Jinyin Chen, Qi~Xuan, and Qingpeng Zhang.
\newblock Unsupervised {{Euclidean Distance Attack}} on {{Network Embedding}}.
\newblock In {\em DSC}, 2020.

\bibitem[\protect\citeauthoryear{Zhu and Ghahramani}{2003}]{zhu2002learning}
Xiaojin Zhu and Zoubin Ghahramani.
\newblock Learning from labeled and unlabeled data with label propagation.
\newblock Technical report, Carneige Mellon University, 07 2003.

\bibitem[\protect\citeauthoryear{Z{\"u}gner and
  G{\"u}nnemann}{2019}]{zugner2019adversarial}
Daniel Z{\"u}gner and Stephan G{\"u}nnemann.
\newblock Adversarial {{Attacks}} on {{Graph Neural Networks}} via {{Meta
  Learning}}.
\newblock In {\em ICLR}, 2019.

\bibitem[\protect\citeauthoryear{Z{\"u}gner \bgroup \em et al.\egroup
  }{2018}]{zugner2018adversarial}
Daniel Z{\"u}gner, Amir Akbarnejad, and Stephan G{\"u}nnemann.
\newblock Adversarial {{Attacks}} on {{Neural Networks}} for {{Graph Data}}.
\newblock In {\em SIGKDD}, 2018.

\end{thebibliography}
}

\newpage
\appendix

\appendix
\section{Appendix}

\subsection{Notations}
The main notations can be found in the Table~\ref{tab:notations}.

\begin{table}[h]
\small
\begin{tabular}{p{1.8cm}p{6cm}}
\toprule
Notation & Description\\ \midrule  \midrule 
$G, A, S, E$ & The original graph, the adjacency matrix, the  generic graph filter and the edge set of $G$ \\
$G^\prime, A^\prime, S^\prime, E^\prime$ & The perturbed graph, the adjacency matrix, the  generic graph filter and the edge set of $G^\prime$ \\
$u, u^\prime$ & The generalized eigenvectors of $A$ and $A^\prime$ \\
$\lambda, \lambda^\prime$ & The generalized eigenvalues of $A$ and $A^\prime$ \\
$\lambda(A), \lambda(A^{\prime})$ & The eigenvalues of $A$ and $A^{\prime}$ \\
$\lambda(S), \lambda(S^{\prime})$ & The eigenvalues of $S$ and $S^{\prime}$ \\
\midrule
$\alpha$ & The normalization parameter of the generic graph filter $S$\\
$\delta$ & The perturbation budget \\
$k$ & The spatial coefficient \\
$\theta$ & The restart threshold\\
$r$ & The number of restart \\
$\epsilon$ & The approximation error of the perturbed eigenvectors \\
\bottomrule
\end{tabular}
\caption{Definition of major symbols.}
\label{tab:notations}
\end{table}

\subsection{Proofs and Derivations} \label{app:proof}

\begin{proof}[Proof of Lemma~\ref{lemma:eigen}]
$Au = \lambda Du$ implies
\[
(D^{-\alpha} AD^{-1+\alpha}) (D^{1-\alpha}u) = \lambda  (D^{1-\alpha}u)
\]
for any real symmetric $A$ and any positive definite $D$, which completes the proof.
\end{proof}

\begin{proof}[Proof of Theorem~\ref{theory:l2}]
	According to the triangle inequality, $\ell_1(A^\prime)$ is lower bounded by
	\begin{linenomath}\small
	\begin{align*}
	    \|(S^\prime)^k - S^k\|_F^2 &\geq (\|(S^\prime)^k\|_F-\|S^k\|_F)^2 \\
	    &= \left(\sqrt{\sum_{i=1}^{n} \lambda_i(S^{\prime})^{2k}}-\sqrt{\sum_{i=1}^{n}  \lambda_i\left(S\right)^{2k}}\right)^{2} \\
	    &= \left(\sqrt{\sum_{i=1}^N (\lambda^\prime_i)^{2k}}-\sqrt{\sum_{i=1}^N \lambda_i^{2k}}\right)^2, 
	\end{align*}
	\end{linenomath}
where the last step follows from Lemma~\ref{lemma:eigen}.
	
	    By applying the Weyl's inequalities (cf. Theorem~\ref{tm:wely} in \S~\ref{sec:addthm})~\cite{horn2008topics} twice, and according to 
        $(d^{\prime}_{1})^{-\alpha} \geq (d^{\prime}_{2})^{-\alpha}  \geq \cdots \geq (d^{\prime}_{N})^{-\alpha} $ and   $(d^{\prime}_{1})^{-1+\alpha} \geq (d^{\prime}_{2})^{-1+\alpha} \geq \cdots \geq (d^{\prime}_{N})^{-1+\alpha}$
    because $\alpha \in [0,1]$, we have
    \begin{linenomath}\small
    \begin{align}
        \lambda_{i}\left(S^{\prime}\right)  
        & = \lambda_{i}\left({D^{\prime}}^{-\alpha} A^{\prime} {D^{\prime}}^{-1+\alpha} \right) \nonumber\\
        & \leq \lambda_{1}\left({D^{\prime}}^{-\alpha}\right) \lambda_{i}\left(A^{\prime} {D^{\prime}}^{-1+\alpha} \right)   \nonumber\\
        &  \leq \lambda_{1}\left({D^{\prime}}^{-\alpha}\right) \lambda_{i}\left(A^{\prime}\right) \lambda_{1}\left( {D^{\prime}}^{-1+\alpha} \right) \nonumber\\
        & = {d^{\prime}_1}^{-\alpha} \lambda_{i}\left(A^{\prime}\right)  {d^{\prime}_1}^{-1+\alpha} \nonumber\\
        & = \frac{1}{d_{\min }^{\prime}} \lambda_{i}\left(A^{\prime}\right),
        \nonumber
     \end{align}
     \end{linenomath}
    where $\lambda_{1}$ is the largest eigenvalue and $d^{\prime}_{1} = d^{\prime}_{\min}$ is the smallest degree in $G^\prime$.
\end{proof}

\begin{proof}[Proof of Theorem 3.]
	According to one iteration of power iteration, when $\lambda_k \neq 0$,
	\begin{linenomath}\small
	\begin{align}
    	u^{\prime}_k &\approx \frac{(D+\Delta D)^{-1}(A+\Delta A) u_k}{ \left\| (D+\Delta D)^{-1}(A+\Delta A) u_k \right\|_{2}} \nonumber \\
    	&= \frac{(D^{-1}A + \Delta C) u_k}{ \left\| (D^{-1}A + \Delta C) u_k\right\|_{2}} \nonumber \\
	    &= \frac{\lambda_{ k} u_k + \Delta C u_k }{ \left\| \lambda_{ k} u_k + \Delta C u_k\right\|_{2}} \label{eq:power-prf} \\
	    &= \frac{\lambda_k u_k + \Delta C u_k}{\sqrt{\lambda_k^2 \transpose{u_k} u_k+2\lambda_k \transpose{u_k} \Delta C u_k + \transpose{u_k} \Delta \transpose{C} \Delta C u_k}} \nonumber \\
	    &\approx \frac{\lambda_k u_k + \Delta C u_k}{|\lambda_k| }
	    = \mathbf{sign} (\lambda_k) u_k + \frac{\Delta C u_k}{|\lambda_k|}. \nonumber 
	\end{align}
	\end{linenomath}
	When $\lambda_k=0$, we can derive from~\eqref{eq:power-prf} that $u_k^\prime \approx (\Delta Cu_k)/(\|\Delta Cu_k\|_2)$.
\end{proof}

\subsection{Additional Theorems}\label{sec:addthm}

\begin{theorem}(Eigenvalue perturbation theory~\cite{stewart1990matrix}).\label{tm:eigenpert}
	Let {\small$\Delta A = A^{\prime} - A$} and {\small$\Delta D = D^{\prime}-D$} be the perturbations of {\small$A$} and {\small$D$}, respectively. Moreover, the number of non-zero entries of {\small$\Delta A$} and {\small$\Delta D$} are assumed to be much smaller than that of  {\small$A$} and {\small$D$}. Let {\small$(\lambda_k, u_k)$} be the {\small$k$}-th pair of generalized eigenvalues and eigenvectors of {\small$A$}, \emph{i.e.}, {\small$Au_k=\lambda_k Du_k$}. We assume that these eigenvectors are properly normalized such that they satisfy {\small$\transpose{u_j} Du_i = 1$} if {\small$i=j$} and {\small$0$} otherwise. Thus, the {\small$k$}-th generalized eigenvalue {\small$\lambda^\prime_k =\lambda_k+\Delta \lambda_k$} can be approximated by
    {\small
    \[
	    \lambda^\prime_k \approx \lambda_k + \transpose{u_k} (\Delta A-\lambda_k \Delta D) u_k.
	\]
	}

	Moreover, when the eigenvalues are distinct, the $k$-th generalized eigenvector $u^\prime_k$ of $A^\prime$ can be approximated by

    {\small
    \[
	    u^\prime_k \approx \left(1 - \frac{1}{2} \transpose{u_k} \Delta D u_k\right) u_k + \sum_{i \neq k} \frac{\transpose{u_{i}} (\Delta A -\lambda_k \Delta D) u_k}{\lambda_ k-\lambda_i} u_i.
	\]
	}
	
\end{theorem}

\begin{theorem}(Bound of eigenvalues of $S^\prime$). \label{tm:eigenbound}
The {\small$i$}-th generalized eigenvalue of {\small$A^\prime$} (i.e., the {\small$i$}-th eigenvalue of {\small$S^\prime$}) is upper bounded by
\begin{equation} \label{eq:bound}\small
    \lambda^\prime_i = \lambda_i(S^{\prime})  \leq \frac{1}{d_{\min }^{\prime}} \cdot \lambda_{i}(A^{\prime}),
\end{equation}
where {\small$d^{\prime}_{\min}$} is the smallest degree in {\small$G^{\prime}$}, {\small$\lambda_i(S^{\prime})$} and  {\small$\lambda_{i}(A^{\prime})$} are the {\small$i$}-th eigenvalue of {\small$S^{\prime}$} and {\small$A^{\prime}$}, respectively. This suggests that the eigenvalues of {\small$S^{\prime}$} are always bounded by the eigenvalues of {\small$A^{\prime}$}.
\end{theorem}

\begin{theorem} (Weyl's inequality for singular values \cite{horn2008topics}).\label{tm:wely}
Let two symmetric matrices $P, Q \in \mathbb{R}^{N \times N}$. Then, for the decreasingly ordered singular values $\sigma$ of $P$, $Q$ and $PQ$,  we have $\sigma_{i+j-1}(PQ) \leq \sigma_{i}(Q) \times \sigma_{j}(Q)$ for any $1 \leq i,j \leq N$ and $i+j \leq N+1$.
\end{theorem}

\subsection{Algorithm} 
Our detailed attack strategy in given in Algorithm~\ref{alg:approx2}.

\begin{algorithm}[h]\small
	\caption{Overall attack strategy via eigensolution approximation with restart.}
	\label{alg:approx2}
	\begin{flushleft}
		\textbf{Input:} Graph $G=(A)$, perturbation budget $\delta$, restart threshold $\theta$. \\
		\textbf{Output:} Modified Graph $G^{\prime} \leftarrow (A^{\prime})$.
	\end{flushleft}
	\begin{algorithmic}[1]
		\STATE $A^{\prime} \leftarrow A $, $D^{\prime} \leftarrow D $;
		\STATE Solve the exact eigensolutions:
		$A^{\prime} u^\prime = \lambda^\prime D^{\prime} u^\prime $;
		\STATE $ \text{Cand} \leftarrow \operatorname{candidate} (A^{\prime}) $;
		\WHILE {$\left \| A^{\prime} - A \right \|_0 \leq 2 \delta$}
		\STATE $e^{\prime} \leftarrow \underset{e \in \text{Cand} }{\operatorname{argmax}} \ \ell_{2}( e )$, which is approximated via Eq.~\ref{eq:eigen-pert-val-2};
		\STATE $A^{\prime}, D^{\prime}  \leftarrow$ insert or remove $e^{\prime}$ to/from $A^{\prime} $;
		\STATE Approximately update eigenvalue $\lambda^{\prime}$ via Eq.~\ref{eq:eigen-pert-val-2} and update eigenvector $u^{\prime}$ via Eq.~\ref{eq:power};
		\IF {$u^{\prime} != \textbf{0}$}
    		\STATE Normalize the perturbed eigenvector $u^{\prime}$ s.t. $\transpose{u^{\prime}_j} Du^{\prime}_i = 1$ if $i=j$;
    		\STATE Compute $\epsilon$ via Eq.~\ref{eq:error};
    		\IF {$\epsilon  > \theta$}
    		 \STATE Recompute the exact eigensolutions;
    		\ENDIF  
		\ELSE
		    \STATE Recompute the exact eigensolutions;
		\ENDIF
		\STATE $\text{Cand} \leftarrow$ remove $e^{\prime}$;
		\ENDWHILE
		\STATE $G^{\prime} \leftarrow (A^{\prime})$.
	\end{algorithmic}
\end{algorithm}

\subsection{Dataset Details}\label{app:dataset}
\vpara{Synthetic datasets.}
We use four synthetic random graphs to evaluate the approximation quality. 
All synthetic graphs have 1000 nodes and parameters are chosen so that the average degree is approximately 10. 
Specifically, for the Erd\H{o}s--R\'enyi graph, we set the probability for edge creation as 0.01.
For the Barab\'asi--Albert graph, we set the number of edges attached from a new node to existing nodes as 5.
When generating the Watts--Strogatz graph, each node is connected to its 10 nearest neighbors in a ring topology, and the probability of rewiring each edge is 0.1.
When generating growing graphs with the power-law degree distribution ~\cite{holme2002growing}, the number of random edges to add for each new node is 5, and the probability of adding a triangle after adding a random edge is 0.1.

\vpara{Real-world datasets.}
We use three social network datasets: Cora-ML, Citeseer and Polblogs.
The former two are citation networks mainly containing machine learning papers. Here, nodes are documents, while edges are the citation links between two documents. Each node has a human-annotated topic as the class label as well as a feature vector. The feature vector is a sparse bag-of-words representation of the document. 
All nodes are labeled to enable differentiation between their topic categories.
Polblogs is a network of weblogs on the subject of US politics. Links between blogs are extracted from crawls of the blog's homepage. The blogs are labelled to identify their political persuasion (liberal or conservative).
Detailed statistics of the social network datasets are listed in Table~\ref{tab:social}.

We also use two protein graph datasets: Proteins and Enzymes.
Proteins is a dataset in which nodes represent secondary structure elements (SSEs) and two nodes are connected by an edge if they are neighbors in either the amino-acid sequence or 3D space.
The label indicates whether or not a protein is a non-enzyme.
Moreover, Enzymes is a dataset of protein tertiary structures.
The task is to correctly assign each enzyme to one of the six EC top-level classes. 
More detailed statistics of the protein graph datasets are listed in  Table~\ref{tab:protein}.

\begin{table}[t]
	\centering
	\small
	\renewcommand\tabcolsep{3.0pt}
\resizebox{0.8\columnwidth}{!}{
        \begin{tabular}{lccc}
        \toprule
        \textbf{dataset} & Cora-ML & Citeseer & Polblogs \\
        \midrule
        \textbf{type} & citation network & citation network & web network \\
        \textbf{\# vertices} & 2,810 & 2,110 & 1,222 \\
        \textbf{\# edges} & 7,981 & 3,757 & 16,714 \\
        \textbf{\# classes} & 7 & 6 & 2 \\
        \textbf{\# features} & 1,433 & 3,703 & 0 \\ \bottomrule
		\end{tabular}
}
		\caption{Data statistics of social datasets.}
		\label{tab:social}
\end{table}

\begin{table}[t]
	\centering
	\small
\resizebox{0.65\columnwidth}{!}{
\renewcommand\tabcolsep{3.0pt}
        \begin{tabular}{lcc}
        \toprule
        \textbf{dataset} & Proteins & Enzymes \\ \midrule
        \textbf{type} & protein network & protein network \\
        \textbf{\# graphs} & 1,113 & 600 \\
        \textbf{\# classes} & 2 & 6 \\
        \textbf{\# features} & 3 & 3 \\
        \textbf{avg \# nodes} & 39.06 & 32.63 \\
        \textbf{avg \# edges} & 72.82 & 62.14 \\ \bottomrule
		\end{tabular}
}
		\caption{Data statistics of protein datasets.}\label{tab:protein}     
\end{table}

\begin{figure*}[h]
    \centering
	\subfloat[Degree centrality]{
		\includegraphics[width=2in]{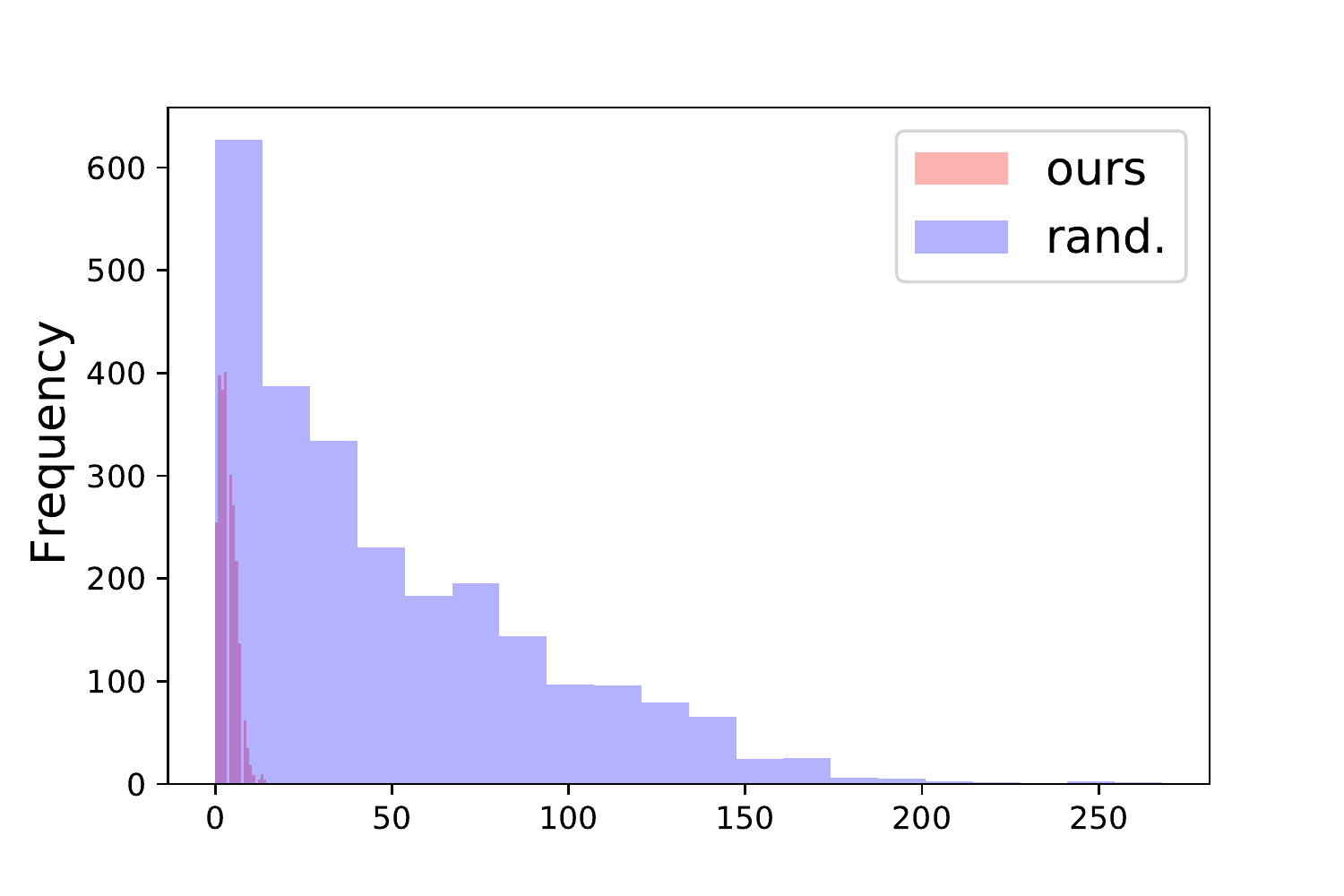}} 
\hspace{0.02\textwidth}
	\subfloat[Betweenness centrality]{
		\includegraphics[width=2in]{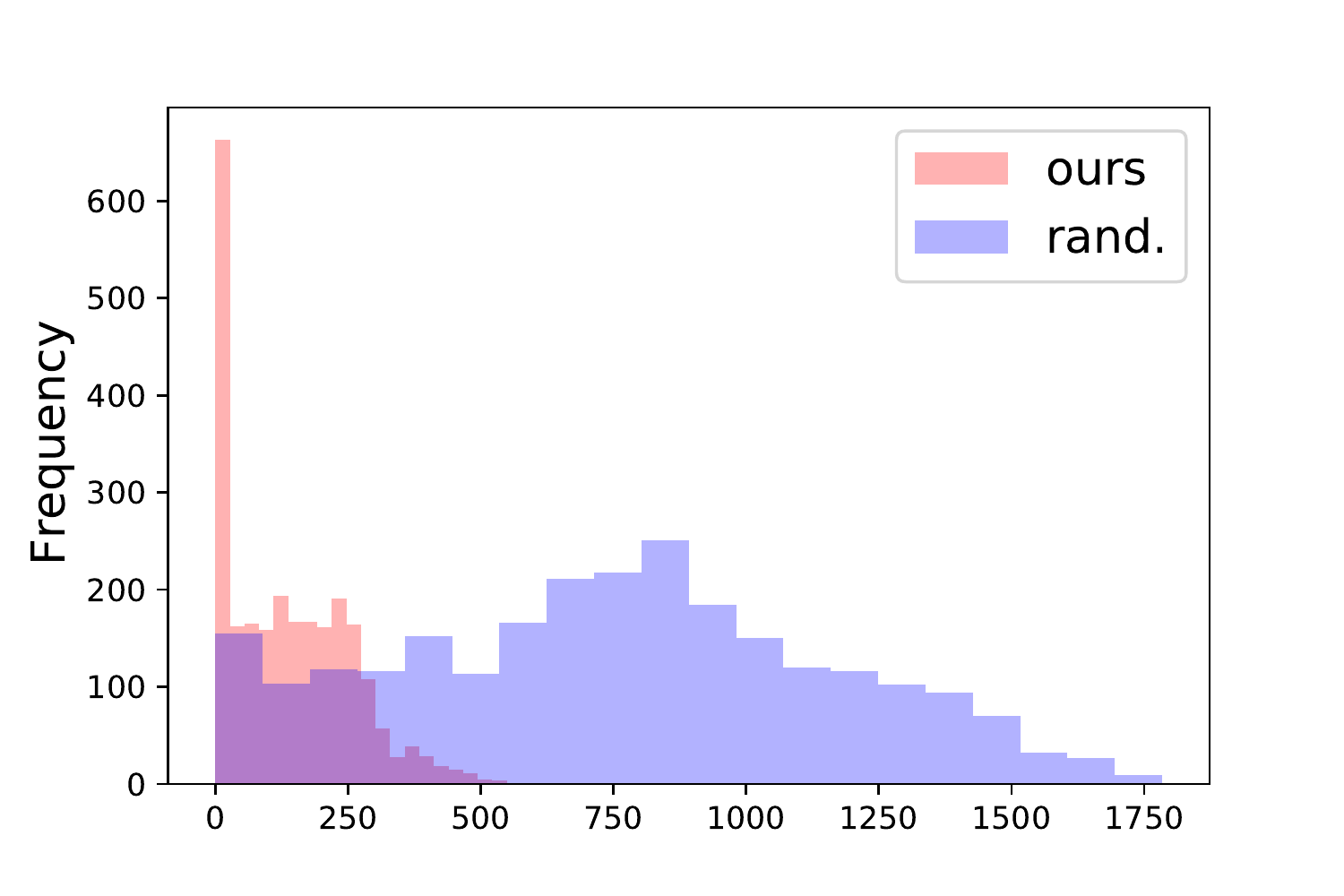}}
\hspace{0.02\textwidth}
	\subfloat[Eigenvector centrality]{
		\includegraphics[width=2in]{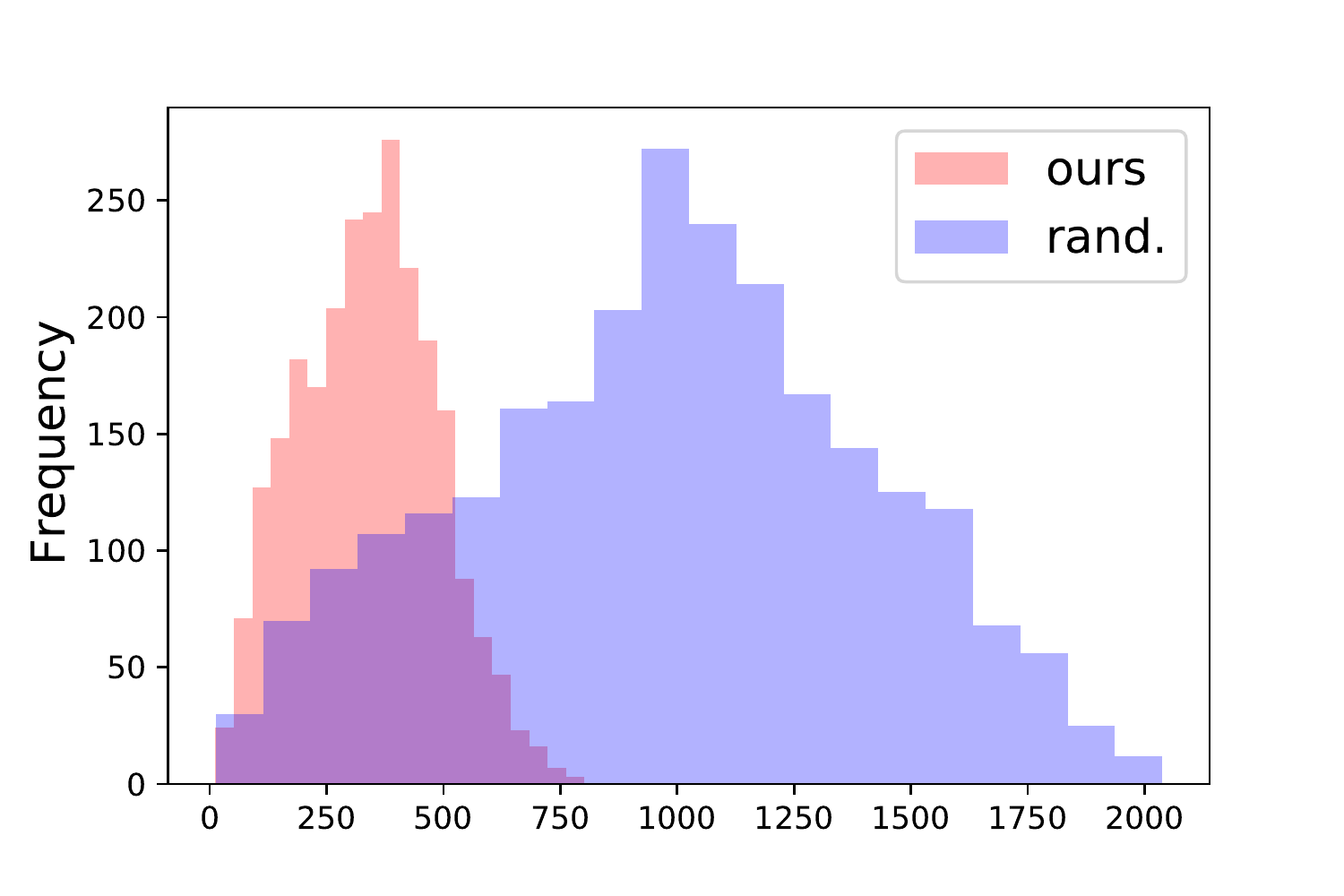}}
	\caption{The comparison of the centrality distributions for our selected adversarial edges and  those of  the randomly selected edges on Polblogs dataset.}
    \label{fig:analyze}
\end{figure*}

\subsection{Implementation Details}\label{app:imp}
Here, we provide additional implementation details of our experiments in \S~\ref{sec:exp}.

For GCN, GIN and Diffpool, we use PyTorch Geometric~\footnote{\url{https://github.com/rusty1s/pytorch_geometric}} for implementation.
We set the learning rate as 0.01 and adopt Adam as our optimizer. For Node2vec, we use its default hyper-parameter setting~\cite{grover2016node2vec}, but set the embedding dimension to 64 and use the implementation by CogDL~\footnote{\url{https://github.com/THUDM/cogdl}}. 
A logistic regression classifier is then used for classification given the embedding. For Label Propagation, we use an implementation that is adapted to graph-structured data~\footnote{\url{https://github.com/thibaudmartinez/label-propagation}}.

For all victim models, we tune the number of epochs based on convergence performance. When performing GCN on Cora-ML, Citeseer and Polblogs, the number of epochs is set to 100. 
When performing GIN or Diffpool on Enzymes, we set the number of epochs to 20, while when performing GIN on Proteins, the number of epochs is set to 10. Finally, when performing Diffpool on Proteins, the number of epochs is set to 2. 
We conduct all experiments on a single machine of Linux system with an Intel Xeon E5 (252GB memory) and a NVIDIA TITAN GPU (12GB memory).

When implementing baselines, we directly apply the default hyperparameters of $\textit{DW}$~\cite{bojchevski2019adversarial}, another black-box attack~\cite{chang2020restricted} and  \textit{Nettack}~\cite{zugner2018adversarial}.
For $\textit{GPGD}$~\cite{xu2019topology} adopting $\ell_1 \left(A^{\prime}\right)$ as the objective, we train 500 epochs with a decaying learning rate as $\text{lr} = \text{lr\_init}(1/(1+i*\text{decay\_rate}))$ where $\text{lr\_init}$ is the initial learning rate, $\text{decay\_rate}$ is the decaying rate and $i$ is the current number of iterations. We set the initial learning rate as 2e-4 with a decaying rate 0.2 on Cora-ML, set the initial learning rate 1e-3 as with a decaying rate 0.2 on Citeseer, Polblogs, and set the initial learning rate 1e-2 as with a decaying rate 0.2 on Proteins and Enzymes.
When applying $\textit{GPGD}$ for the white-box setting, we set the learning rate as 10 and run 700 epochs.

\subsection{Additional Experimental Results} \label{app:exp}

\vpara{Node-level attack with increasing perturbation rate.}
Table~\ref{tab:pert_rate} further shows that under increasing perturbation rates (5/10/15\%), our node-level attacker can do more damage to GCN while still achieves the best performance. Note that when the perturbation rate is not very high, our solution without restart is already good enough to mount attacks. 

\begin{table}[h]
\centering
\small
\resizebox{1\columnwidth}{!}{
	\renewcommand\tabcolsep{2.5pt} 
    \begin{tabular}{*4{c}|*4{c}}
    \toprule
    \diagbox [width=7em,height=1.7\line,trim=l] {attacker}{pert. rate} 
     & 5\% & 10\% & 15\% &\diagbox [width=7em,height=1.7\line,trim=l] {attacker}{pert. rate} & 5\% & 10\% & 15\%  \\ \hline
    \textit{Rand.} & 0.75 & 1.97 & 2.41 & \textit{GPGD} & 3.94 & 4.22 & 5.03 \\ 
    \textit{Deg.} & 0.59 & 1.12 & 1.34 & \textit{GF-Attack} &1.10 &1.34 &2.11 \\ 
    \textit{Betw.} & 0.63 & 1.22 & 1.45 &
    \textit{\ours-r-d} &3.90 &4.03 &5.11 \\ 
    \textit{Eigen.} & 0.30 & 0.28 & 1.12 & \textit{\ours-r} & \textbf{4.43} & 5.02 & 5.77 \\ 
    \textit{DW} & 0.34 & 0.85 & 1.23 &   \textit{\ours} & 4.30 & \textbf{5.27} & \textbf{6.40}
  \\ 
    \bottomrule 
    \end{tabular}
}
\caption{Decrease in Macro-F1 score with different perturbation rates when attacking GCN on Cora-ML.}
\label{tab:pert_rate}
\end{table}

\vpara{Attacks on Defensive models.}
We further validate the effectiveness of our attacks against three defensive models: EdgeDrop~\cite{dai2018adversarial}, Jaccard~\cite{wu2019adversarial} and SVD~\cite{entezari2020all}. We utilize {GCN} as the backbone classifier and test on node classification task. 
We compare our method with the strongest competitor \textit{GPGD}.
The experimental settings are the same as those used in node-level attack.
The results in Table~\ref{tab:defense} demonstrate that our method outperforms \textit{GPGD} in terms of its ability to attack the defensive models. We can see that EdgeDrop cannot effectively defend against
our attack, while the two pre-processing defense approaches (Jaccard and SVD) can defend against our attack to a certain extent. 

\begin{table}[h]
\centering
\renewcommand\tabcolsep{3.0pt} 
\small
\resizebox{0.7\columnwidth}{!}{
\begin{tabular}{ccccc}
\toprule
& w/o defense & EdgeDrop & Jaccard & SVD \\ \midrule 
\textit{GPGD} &4.22 &6.04 &3.94 &3.47\\
\textit{\ours} &5.27 &7.11 &4.71  & 4.02 \\ 
\bottomrule
\end{tabular}
}
\caption{Attack against defensive models on Cora-ML. The decrease in Macro-F1 score is reported here.}
\label{tab:defense}
\end{table}

\begin{table}[b]
\setlength{\tabcolsep}{8pt}
\vspace{-1em}
\centering
\small
\resizebox{0.8\columnwidth}{!}{
\begin{tabular}{ccccc} 
\toprule
  & & $\ell^*_1$ &$\ell^*_2$  & \textit{\ours} 
 \\ \midrule 
\multirow{3}{*}{Cora-ML} & {GCN} &1.92 &1.72 & \textbf{5.27} \\
 & {Node2vec} &8.05 &8.14 & \textbf{8.29} \\
 & {Label Prop.} &5.32 &5.37 & \textbf{7.13} \\ \hline
\multirow{3}{*}{Citeseer} & {GCN} &1.65 &2.27 & \textbf{3.98} \\
 &{Node2vec} &8.64 &9.05 & \textbf{9.32} \\
 & {Label Prop.} &7.09 &5.41 & \textbf{8.16} \\ \hline
\multirow{3}{*}{Polblogs} & {GCN} & 2.65 &2.92 & \textbf{5.32} \\
 & {Node2vec} &3.68 &3.43 & \textbf{3.79} \\
 & {Label Prop.} &4.80 &4.20 & \textbf{6.14} \\ \bottomrule
\end{tabular}
}
\caption{The experimental results of other objective functions of node-level attacks against three types of victim models. We report the decrease in Macro-F1 score (in percent) on the test set after the attack. }
\label{tab:cand}
\end{table}

\begin{table*}[h]
\small
\centering
\renewcommand\tabcolsep{2pt} 
\resizebox{2\columnwidth}{!}{
\begin{tabular}{cccc|cccccccccc|c} 
\toprule
 &  &           & (Unattacked) & \textit{Rand.}        & \textit{Deg.}    & \textit{Betw.} & \textit{Eigen.} & \textit{DW}     &\textit{GF-Attack}  & \textit{GPGD}        &\textit{\ours-r-d} & \textit{\ours-r}       & \textit{\ours}                  & White-box     \\ \midrule
\multirow{9}{*}{Cora-ML}  & \multirow{3}{*}{GCN}         & F1        & 0.82$\pm$0.8 & 1.97$\pm$0.8          & 1.12$\pm$0.4     & 1.22$\pm$0.4   & 0.28$\pm$0.3    & 0.85$\pm$0.3   &1.34$\pm$0.5  & 4.22$\pm$0.6    &4.03$\pm$0.6  & 5.02$\pm$0.4 & \textbf{5.27}$\pm$0.3 & 11.36$\pm$0.5 \\
                          &                              & Prec. & 0.81$\pm$0.9 & 0.55$\pm$0.5          & 0.20$\pm$0.4     & 0.11$\pm$0.3   & 0.18$\pm$0.3    & 0.32$\pm$0.2     &0.41$\pm$0.5 & 1.69$\pm$0.6 &1.44$\pm$0.6 & 3.02$\pm$0.4 & \textbf{3.43}$\pm$0.3 & 6.58$\pm$0.6  \\
                          &                              & Recall    & 0.82$\pm$1.1 & 1.15$\pm$0.9          & 0.94$\pm$0.4     & 1.23$\pm$0.6   & 0.49$\pm$0.4    & 0.87$\pm$0.4   &1.50$\pm$0.6  & 2.51$\pm$0.6  &2.10$\pm$0.6 &3.84$\pm$0.4 & \textbf{4.05}$\pm$0.4 & 6.87$\pm$0.4  \\
                          \cdashline{2-13}
                          & \multirow{3}{*}{Node2vec}    & F1        & 0.79$\pm$0.8 & 6.37$\pm$1.8          & 5.40$\pm$1.6     & 3.33$\pm$1.0   & 2.84$\pm$1.0    & 3.25$\pm$1.3  &5.76$\pm$1.5    & 5.33$\pm$1.8  &5.82$\pm$1.7 & 6.92$\pm$1.0 & \textbf{8.29}$\pm$1.0 & (1.43$\pm$0.9)  \\
                          &                              & Prec. & 0.78$\pm$0.7 & 6.06$\pm$1.8          & 3.04$\pm$1.7     & 3.64$\pm$0.9   & 2.15$\pm$1.1    & 2.96$\pm$1.3   &5.03$\pm$1.4  & 4.96$\pm$1.7  &5.21$\pm$1.6 & 5.95$\pm$1.0 & \textbf{8.04}$\pm$1.0 & (1.39$\pm$1.0)  \\
                          &                              & Recall    & 0.78$\pm$0.9 & 6.69$\pm$2.0          & 3.76$\pm$1.5     & 4.92$\pm$1.3   & 3.01$\pm$1.0    & 3.57$\pm$1.4  &5.22$\pm$1.6    & 5.10$\pm$1.8  &5.30$\pm$1.7 & 6.32$\pm$1.0 & \textbf{8.54}$\pm$0.9 & (1.52$\pm$1.1)  \\
                          \cdashline{2-13}
                          & \multirow{3}{*}{Label Prop.} & F1        & 0.80$\pm$0.7 & 4.10$\pm$1.3          & 2.45$\pm$0.7     & 2.71$\pm$0.8   & 2.07$\pm$0.7    & 1.79$\pm$0.9    &3.18$\pm$0.5 & 4.28$\pm$1.6   &5.01$\pm$0.7  & 6.02$\pm$0.9 & \textbf{7.13}$\pm$0.9 & (1.05$\pm$1.0)  \\
                          &                              & Prec. & 0.81$\pm$0.8 & 2.98$\pm$1.3          & 1.10$\pm$0.6     & 1.20$\pm$0.4   & 1.46$\pm$0.7    & 1.05$\pm$0.8     &2.74$\pm$0.4 & 3.05$\pm$1.0 &4.10$\pm$0.6  & 4.62$\pm$0.8 & \textbf{4.98}$\pm$1.0 & (0.09$\pm$0.8)  \\
                          &                              & Recall    & 0.80$\pm$1.0 & 4.95$\pm$1.2          & 3.61$\pm$0.9     & 4.32$\pm$1.0   & 2.64$\pm$0.9    & 2.54$\pm$1.2  &4.71$\pm$0.6    & 5.03$\pm$1.9  &5.15$\pm$0.7 & 7.92$\pm$1.0 & \textbf{7.99}$\pm$1.0 & (1.51$\pm$1.0)  \\\midrule
\multirow{9}{*}{Citeseer} & \multirow{3}{*}{GCN}         & F1        & 0.66$\pm$1.4 & 2.02$\pm$0.6          & 0.16$\pm$0.4     & 0.70$\pm$0.4   & 0.64$\pm$0.4    & 0.21$\pm$0.4   &1.36$\pm$0.7   & 2.14$\pm$0.9  &2.63$\pm$0.7 & 3.16$\pm$0.6 & \textbf{3.98}$\pm$0.5 & 6.42$\pm$0.6  \\
                          &                              & Prec.           & 0.64$\pm$1.6 & 1.15$\pm$0.7          & 0.10$\pm$0.5     & 0.33$\pm$0.2   & 0.30$\pm$0.4    & 0.09$\pm$0.5   &1.01$\pm$0.6  & 2.01$\pm$0.8   &2.50$\pm$0.7 & 2.94$\pm$0.6 & \textbf{3.28}$\pm$0.7 & 5.92$\pm$0.6  \\
                          &                              &Recall           & 0.64$\pm$1.6 & 1.90$\pm$0.6          & -81.41$\pm$245.3 & 0.72$\pm$0.4   & 0.64$\pm$0.5    & 0.13$\pm$0.4  &1.16$\pm$0.7   & 2.51$\pm$0.9   &2.66$\pm$0.8 & 3.62$\pm$0.7 & \textbf{4.00}$\pm$0.5 & 7.06$\pm$0.5  \\
                          \cdashline{2-13}
                          & \multirow{3}{*}{Node2vec}    & F1        & 0.60$\pm$1.5 & 7.47$\pm$2.3          & 7.47$\pm$1.6     & 3.47$\pm$2.6   & 4.87$\pm$1.5    & 2.54$\pm$2.5  &6.45$\pm$3.5   & 5.26$\pm$1.9   &7.94$\pm$1.6  & 8.32$\pm$2.5 & \textbf{9.32}$\pm$2.6 & (0.12$\pm$1.0)  \\
                          &                              &  Prec.         & 0.59$\pm$1.6 & 7.28$\pm$2.4          & 7.33$\pm$2.1     & 3.01$\pm$2.7   & 4.26$\pm$1.5    & 2.88$\pm$3.2    &6.21$\pm$3.0 & 5.00$\pm$1.9  &7.56$\pm$1.5  & 7.91$\pm$2.4 & \textbf{8.69}$\pm$2.7 & (0.10$\pm$1.0)  \\
                          &                              &   Recall        & 0.62$\pm$1.5 & 7.24$\pm$2.2          & 6.94$\pm$1.6     & 4.51$\pm$2.4   & 5.00$\pm$1.4    & 2.62$\pm$2.5    &6.51$\pm$3.7  & 5.86$\pm$1.9 &8.06$\pm$1.6  & 8.69$\pm$2.5 & \textbf{9.86}$\pm$2.7 & (0.19$\pm$1.0)  \\
                          \cdashline{2-13}
                          & \multirow{3}{*}{Label Prop.} & F1        & 0.64$\pm$0.8 & 6.70$\pm$2.0          & 3.47$\pm$0.8     & 6.00$\pm$1.7   & 5.36$\pm$0.6    & 3.00$\pm$0.  &6.99$\pm$1.0  & 5.14$\pm$1.9  &6.66$\pm$1.3 & 7.79$\pm$0.9 & \textbf{8.16}$\pm$0.9 & (2.47$\pm$1.2)  \\
                          &                              & Prec. & 0.62$\pm$0.8 & 6.01$\pm$1.9          & 2.95$\pm$0.6     & 5.59$\pm$1.8   & 5.06$\pm$0.6    & 2.54$\pm$0.6   &4.03$\pm$0.8  & 4.52$\pm$1.8  &6.23$\pm$1.3  & 6.69$\pm$1.0 & \textbf{7.77}$\pm$0.9 & (2.06$\pm$1.1)  \\
                          &                              & Recall    & 0.67$\pm$0.7 & 6.88$\pm$2.0          & 3.01$\pm$0.9     & 4.35$\pm$1.8   & 6.02$\pm$0.6    & 3.10$\pm$0.9    &7.03$\pm$1.0  & 5.45$\pm$1.9  &7.25$\pm$1.4& 8.16$\pm$0.9 & \textbf{8.55}$\pm$1.0 & (2.66$\pm$1.2)  \\\midrule
\multirow{9}{*}{Polblogs} & \multirow{3}{*}{GCN}         & F1        & 0.96$\pm$0.7 & 1.91$\pm$1.5          & 0.03$\pm$0.2     & 1.72$\pm$0.6   & 0.67$\pm$0.5    & 0.01$\pm$0.4     &1.15$\pm$0.4  & 2.35$\pm$1.8 &3.06$\pm$1.2 & 4.30$\pm$1.2 & \textbf{5.32}$\pm$1.1 & 3.88$\pm$1.1  \\
                          &                              & Prec.          & 0.95$\pm$0.7 & 1.65$\pm$1.0          & 0.02$\pm$0.2     & 1.51$\pm$0.5   & 0.35$\pm$0.5    & 0.02$\pm$0.3     &0.98$\pm$0.3  & 2.10$\pm$1.8  &2.86$\pm$1.2 & 4.02$\pm$1.1 & \textbf{4.99}$\pm$1.0 & 3.05$\pm$1.0  \\
                          &                              & Recall          & 0.96$\pm$0.7 & 1.93$\pm$1.5          & 0.01$\pm$0.1     & 1.74$\pm$0.6   & 0.75$\pm$0.5    & 0.01$\pm$0.4   &1.17$\pm$0.3    & 2.53$\pm$1.9 &3.36$\pm$1.0  & 4.58$\pm$1.2 & \textbf{6.01}$\pm$1.0 & (4.06$\pm$1.1)  \\
                          \cdashline{2-13}
                          & \multirow{3}{*}{Node2vec}    & F1        & 0.95$\pm$0.3 & 3.01$\pm$0.7 & 0.04$\pm$0.6     & 3.07$\pm$0.6   & 1.84$\pm$0.3    & 0.18$\pm$0.4   &1.00$\pm$0.5    & 2.49$\pm$0.6 &2.57$\pm$0.9 & 2.74$\pm$0.5 & \textbf{3.79}$\pm$0.5          & (2.13$\pm$0.4)  \\
                          &                              &Prec.           & 0.95$\pm$0.3 & 2.01$\pm$0.6 & 0.03$\pm$0.5     & 3.06$\pm$0.6   & 1.22$\pm$0.4    & 0.17$\pm$0.5   &0.86$\pm$0.5  & 2.06$\pm$0.7  &2.11$\pm$0.8  & 2.65$\pm$0.5  & \textbf{3.51}$\pm$0.4          & (2.09$\pm$0.4)  \\
                          &                              &Recall           & 0.95$\pm$0.3 & 3.04$\pm$0.7 & 0.04$\pm$0.4     & 3.04$\pm$0.6   & 2.66$\pm$0.3    & 0.18$\pm$0.8   &1.12$\pm$0.5   & 2.66$\pm$0.5 &2.63$\pm$0.9  & 2.94$\pm$0.5 & \textbf{3.99}$\pm$0.5          & (2.13$\pm$0.3)  \\
                          \cdashline{2-13}
                          & \multirow{3}{*}{Label Prop.} & F1        & 0.96$\pm$0.5 & 4.99$\pm$0.7          & 0.08$\pm$0.4     & 3.45$\pm$0.7   & 2.15$\pm$0.3    & 0.37$\pm$0.5   &2.18$\pm$0.4    & 4.15$\pm$0.8 &5.17$\pm$0.8 & 5.84$\pm$0.7 & \textbf{6.14}$\pm$0.7 & (2.28$\pm$0.5)  \\
                          &                              & Prec. & 0.96$\pm$0.5 & 4.41$\pm$0.8          & 0.10$\pm$0.4     & 3.03$\pm$0.6   & 2.45$\pm$0.8    & 0.34$\pm$0.4      &1.99$\pm$0.4 & 3.51$\pm$0.7  &4.91$\pm$0.8 & 5.05$\pm$0.8 & \textbf{5.08}$\pm$0.7 & (2.05$\pm$0.5)  \\
                          &                              & Recall    & 0.96$\pm$0.5 & 4.67$\pm$0.8          & 0.04$\pm$0.4     & 3.15$\pm$0.7   & 2.05$\pm$0.8    & -87.62$\pm$264.2  &2.20$\pm$0.4 & 4.22$\pm$0.8 &5.30$\pm$0.9 & 5.86$\pm$0.7 & \textbf{6.56}$\pm$0.6 & (2.51$\pm$0.5) \\
\bottomrule
\end{tabular}
}
\caption{The detailed experimental results with standard deviation of node-level attacks against three types of victim models. We report the decrease of performance (in percent) on the test set after the attack is performed; the higher the better (note that negative results denote that the performance adversely increases after the attack). We also report the performance on the unattacked graph.
}
\label{tab:detail}
\end{table*}

c

\vpara{Can heuristics explain adversarial edges?} 
A most straightforward strategy of identifying adversarial edges is to utilize simple heuristics to capture ``important'' edges~\cite{bojchevski2019adversarial}. However, recall that in Table~\ref{tab:node_attack}, some results unexpectedly revealed that some heuristic methods (\emph{i.e.}, \textit{Deg.}, \textit{Betw.}, \textit{Eigen.}) sometimes performs worse than randomly selection of adversarial edges. We here analyze this observation by comparing the degree, betweenness or eigenvector centrality distribution of our selected adversarial edges with that of the randomly selected edges. In Figure~\ref{fig:analyze}, we find that our selected adversarial edges tend to have smaller degree, betweenness or eigenvector centrality. 
Whereas, common heuristic methods, following a previous work~\cite{bojchevski2019adversarial}, are performed by selecting the adversarial edges with bigger centrality (indicating their importance).
Our findings actually run counter to these heuristic methods, thus this is one possible reason why they perform badly. 
Moreover, Eq.~\ref{eq:bound} gives us an intuitive spectral view to analyze the degree centrality, \emph{i.e.}, the spectrum of $S^\prime$ is upper bounded by a term w.r.t the smallest degree in $G^\prime$.

We therefore propose to select adversarial edges based on the increasing order of the sum of the degree/betweenness/eigenvector centrality (\emph{i.e.}, \textit{SmallDeg.}, \textit{SmallBetw.} and \textit{SmallEigen.}) and report their results as the setting of node-level attack in Table~\ref{tab:heur-small}. We observe that the smaller centrality does perform better than the larger one in most cases although still can not beat our method.

\begin{table}[htbp]
\centering 
\small
\renewcommand\tabcolsep{3pt} 
\resizebox{0.8\columnwidth}{!}{
\begin{tabular}{ccccc}
\toprule
\multicolumn{1}{l}{} &  & \textit{SmallDeg.} & \textit{SmallBetw.} & \textit{SmallEigen.} \\ \midrule
\multirow{3}{*}{Cora-ML} & GCN & 2.27 $\pm$ 0.4 & 2.29 $\pm$ 0.6 & 1.44 $\pm$ 0.4 \\
 & node2vec & 4.47 $\pm$ 0.9 & 5.64 $\pm$ 1.4 & 4.85 $\pm$ 0.8 \\
 & Label Prop. & 5.14 $\pm$ 0.6 & 5.37 $\pm$ 0.5 & 4.63 $\pm$ 0.3 \\ \midrule
\multirow{3}{*}{Citeseer} & GCN & 2.77 $\pm$ 0.7 & 2.81 $\pm$ 0.6 & 2.34 $\pm$ 0.8 \\
 & node2vec & 5.78 $\pm$ 2.4 & 7.99 $\pm$ 2.0 & 2.26 $\pm$ 1.7 \\
 & Label Prop. & 7.66 $\pm$ 0.8 & 7.53 $\pm$ 1.3 & 5.33 $\pm$ 0.5 \\ \midrule
\multirow{3}{*}{Polblogs} & GCN & 3.14 $\pm$ 0.8 & 2.64 $\pm$ 1.0 & 3.58 $\pm$ 0.7 \\
 & node2vec & 2.00 $\pm$ 0.5 & 1.33 $\pm$ 0.6 & 1.75 $\pm$ 0.7 \\
 & Label Prop. & 4.81 $\pm$ 0.6 & 4.71 $\pm$ 0.7 & 5.92 $\pm$ 0.8 \\ \bottomrule
\end{tabular}
}
\caption{The relative decrease in Macro-F1 score of additional heuristic methods with standard deviation of node-level attacks against three types of victim models.}
\label{tab:heur-small}
\end{table}

\begin{figure}[]
	\centering
	\includegraphics[width=0.3\textwidth]{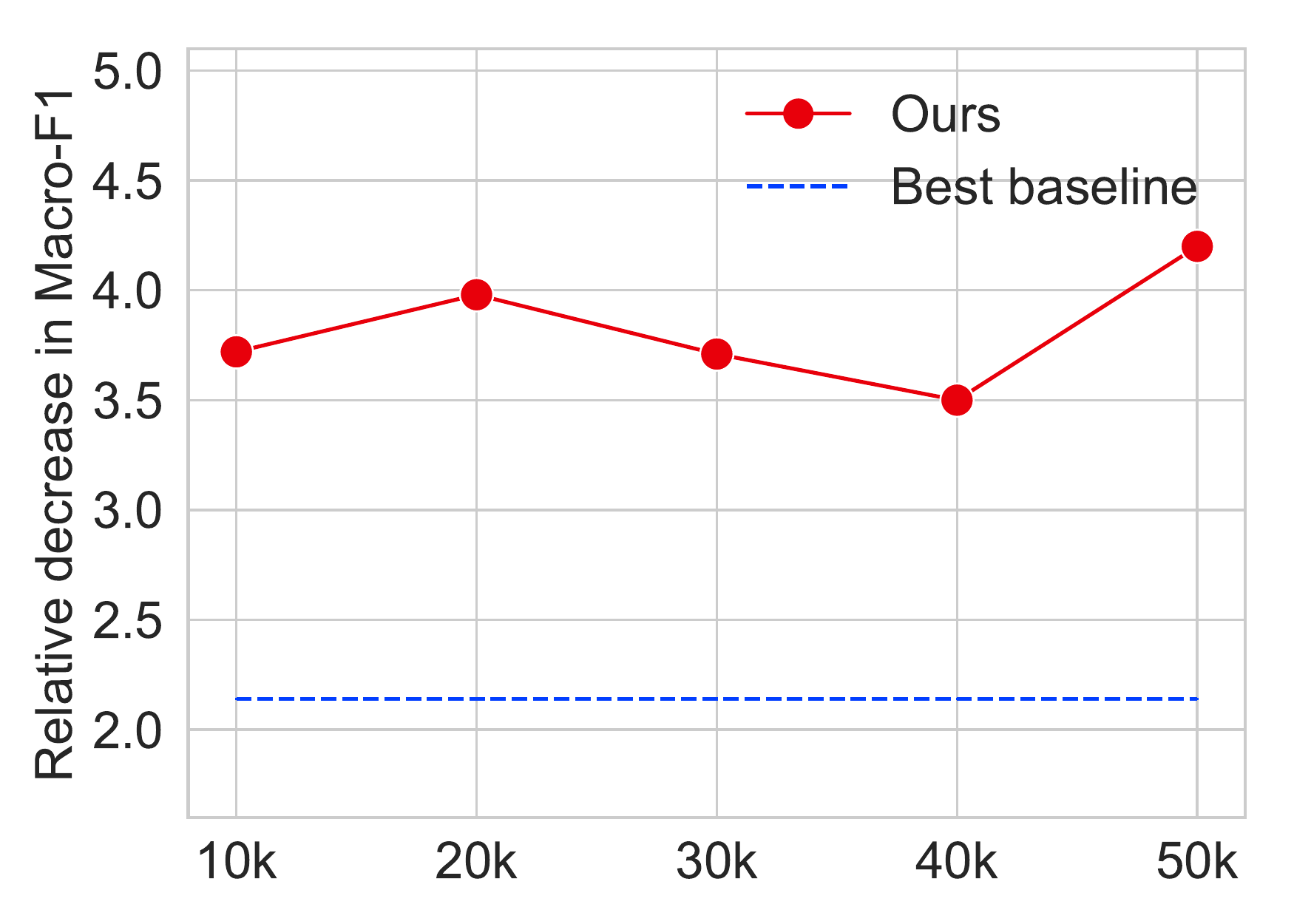}
	\caption{ The relative decrease in Macro-F1 score of node-level attacks against GCN with different size of the candidate set on Citeseer dataset. The blue dashed line denotes the performance of the best baseline when candidate size is chosen as $20\text{k}$. }
	\label{fig:cand}
\end{figure}

\vpara{The selection of candidate set.}
We have mentioned that one limitation of our attack strategy is  the randomly selected candidate set, which introduces a further approximation. We here further analyze the impact of the candidate set size. Figure~\ref{fig:cand} shows the relative decrease when we choose the size of candidate set among $\{10\text{k}, 20\text{k}, 30\text{k} , 40\text{k}, 50\text{k}\}$. We observe that although our model performs slightly better if we randomly select $50\text{k}$ edge pairs as the candidate, the candidate size appears to have little influence on the overall performance, which 
partly verifies the practical utility of our strategy of randomly selecting candidates.

\begin{figure}[H]
    \centering
	\subfloat[$k$: spatial coefficient]{
		\includegraphics[width=1.4in]{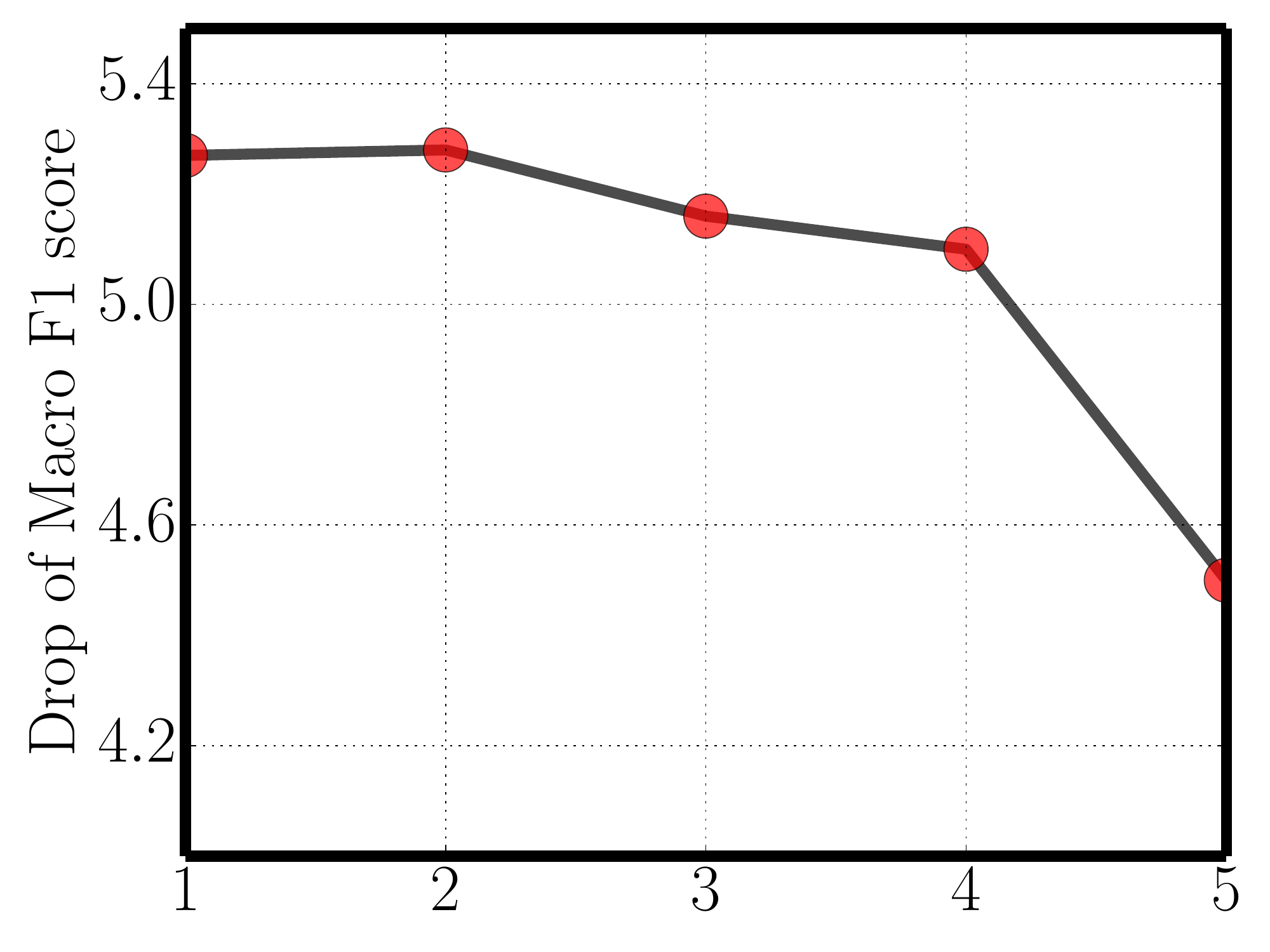}} 
\hspace{0.02\textwidth}
	\subfloat[$\theta$: restart threshold]{
		\includegraphics[width=1.4in]{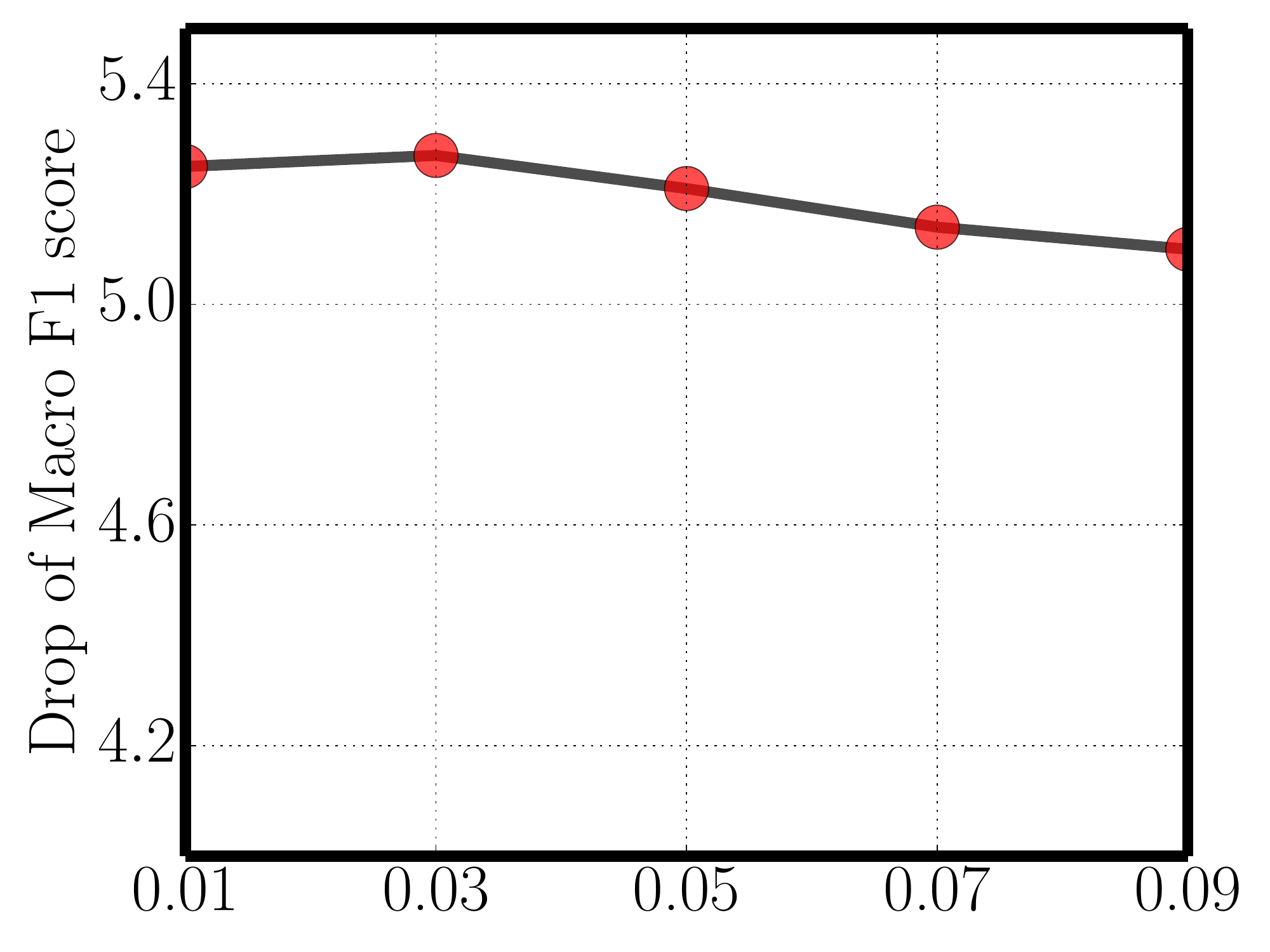}}
	\caption{Parameter analysis for node classification task.}
	\label{fig:parameter}
\end{figure}

\vpara{Parameter Analysis.}
Our algorithm involves two hyperparameters: namely, the spatial coefficient $k$ and restart threshold $\theta$. We use grid search to find their suitable values on Cora-ML and present here for reference (see Figure~\ref{fig:parameter}). 
Better performance can be obtained when $k \in \{1,2\}$ and $\theta \leq 0.03$. The observation that our model's superiority when $k \leq 2$ is consistent with many conclusions of graph over-smoothing problem~\cite{smooth1,smooth2}. Remarkably, our model can still achieve relatively good performance (\emph{i.e.}, better than the baseline methods) regardless of how hyperparameters are changed.

\vpara{Detailed results of Table~\ref{tab:node_attack}.} We further report the detailed results (decrease in Macro-F1 score, Macro-Precision score and Macro-Recall score) with standard deviation of node-level attacks in Table~\ref{tab:detail}.

\hide{
\begin{figure*}[b]  
	\centering  
	\includegraphics[width=0.95\textwidth]{model}\\  
	\caption{An illustrative example of the proposed attack procedure on graphs, in which the budget constraint $\delta$ is set to be 2.}
	\label{fig:model}     
\end{figure*} 
}

\end{document}